\documentclass{article}

     \PassOptionsToPackage{numbers, compress}{natbib}
\usepackage[final]{neurips_2020}

\usepackage[utf8]{inputenc} % allow utf-8 input
\usepackage[T1]{fontenc}    % use 8-bit T1 fonts
\usepackage{hyperref}       % hyperlinks
\usepackage{url}            % simple URL typesetting
\usepackage{booktabs}       % professional-quality tables
\usepackage{amsfonts}       % blackboard math symbols
\usepackage{nicefrac}       % compact symbols for 1/2, etc.
\usepackage{microtype}      % microtypography

\usepackage{graphicx}
\usepackage{multirow}
\usepackage{algorithmic}
\usepackage{algorithm}
\usepackage{paralist,algorithmic,algorithm}
\usepackage{amsmath}
\usepackage{booktabs}
\usepackage{wrapfig}

\newcommand{\x}{{\bf x}}

\newcommand{\y}{{\bf y}}

\newcommand{\D}{\mathcal{D}}

\newcommand{\X}{\mathcal{X}}
\newcommand{\U}{\mathcal{U}}

\newcommand{\R}{\mathcal{R}}

\newtheorem{defn}{Definition}

\title{S2OSC: A Holistic Semi-Supervised Approach for Open Set Classification}

\author{%
  Yang Yang
  \thanks{Yang Yang and Jian Yang are with the Nanjing University of Science and Technology, Nanjing 210094, China.} \\
  Nanjing University of Science and Technology\\
  \texttt{yyang@njust.edu.cn} \\
   \And
   Zhen-Qiang Sun \\
   Nanjing Normal University \\
   \texttt{enderman19980125@outlook.com} \\
   \AND
   Hui Xiong \\
   Rutgers University \\
   \texttt{hxiong@rutgers.edu} \\
   \And
   Jian Yang \\
   Nanjing University of Science and Technology \\
   \texttt{csjyang@njust.edu.cn} \\
}

\begin{document}

\maketitle

\begin{abstract}
Open set classification (OSC) tackles the problem of determining whether the data are in-class or out-of-class during inference, when only provided with a set of in-class examples at training time. Traditional OSC methods usually train discriminative or generative models with in-class data, then utilize the pre-trained models to classify test data directly. However, these methods always suffer from embedding confusion problem, i.e., partial out-of-class instances are mixed with in-class ones of similar semantics, making it difficult to classify. To solve this problem, we unify semi-supervised learning to develop a novel OSC algorithm, S2OSC, that incorporates out-of-class instances filtering and model re-training in a transductive manner. In detail, given a pool of newly coming test data, S2OSC firstly filters distinct out-of-class instances using the pre-trained model, and annotates super-class for them. Then, S2OSC trains a holistic classification model by combing in-class and out-of-class labeled data and remaining unlabeled test data in semi-supervised paradigm, which also integrates pre-trained model for knowledge distillation to further separate mixed instances. Despite its simplicity, the experimental results show that S2OSC achieves state-of-the-art performance across a variety of OSC tasks, including $85.4\%$ of F1 on CIFAR-10 with only 300 pseudo-labels. We also demonstrate how S2OSC can be expanded to incremental OSC setting effectively with streaming data. 
\end{abstract}
%Code is available at: \url{https://github.com/data-ming-and-application/S2OSC/}

\section{Introduction}\label{sec:s2}
The real-world is changing dynamically, and many applications are non-stationary, which always receive the data containing out-of-class (also called unknown class) instances, for example, self-driving cars need to identify unknown objects, face recognition system needs to distinguish unseen personal pictures, image retrieval often emerges new categories, etc. This problem is defined as ``Open Set Classification (OSC)'' in literature~\cite{geng2020}. Different from traditional Closed Set Classification (CSC) which assumes training and testing data are draw from same spaces, i.e., the label and feature spaces, OSC aims to not only accurately classify in-class (also called known class) instances, but also effectively detect out-of-class instances. Besides, a generalized situation is that out-of-class instances will arise continuously with the streaming data, i.e., unknown classes appear incrementally, which is also defined as incremental OSC. 
 
Both anomaly detection~\cite{LiuTZ08,XiaCWHS15} and zero-shot learning~\cite{ChangpinyoCGS16,LiJLZYH19} are related to open set classification. They have similar objectives to detect anomaly/out-of-class instances given a set of in-class examples. In contrast, anomaly detection (also called outlier detection) is an unsupervised learning task~\cite{XiaCWHS15}. The goal is to separate abnormal in-class instances from normal ones, where the distinction from OSC is that differences between unknown and known classes are larger than that between anomalies and known classes~\cite{CaiZTM019}. Unlike anomaly detection, zero-shot learning (ZSL) focuses on constructing related OSC models to address unknown class detection issue, which merely utilize in-class examples and semantic information about unknown classes. Whereas the standard ZSL methods only test out-of-class instances, rather than test both known and unknown classes. Thus, generalized zero-shot learning (GZSL) is proposed, which automatically detect known and unknown classes simultaneously. For example, \cite{ChangpinyoCGS16,LiJLZYH19} learned more reliable classification models by measuring the distance between examples and corresponding in-class/out-of-class semantic embeddings. However, both ZSL and GZSL assume that semantic information (for example, attributes or descriptions) of the unknown classes is given, which is limited to classify with prior knowledge.

\begin{figure}[htb]
	\begin{center}
		\begin{minipage}[h]{33mm}
			\centering
			\includegraphics[width=33mm]{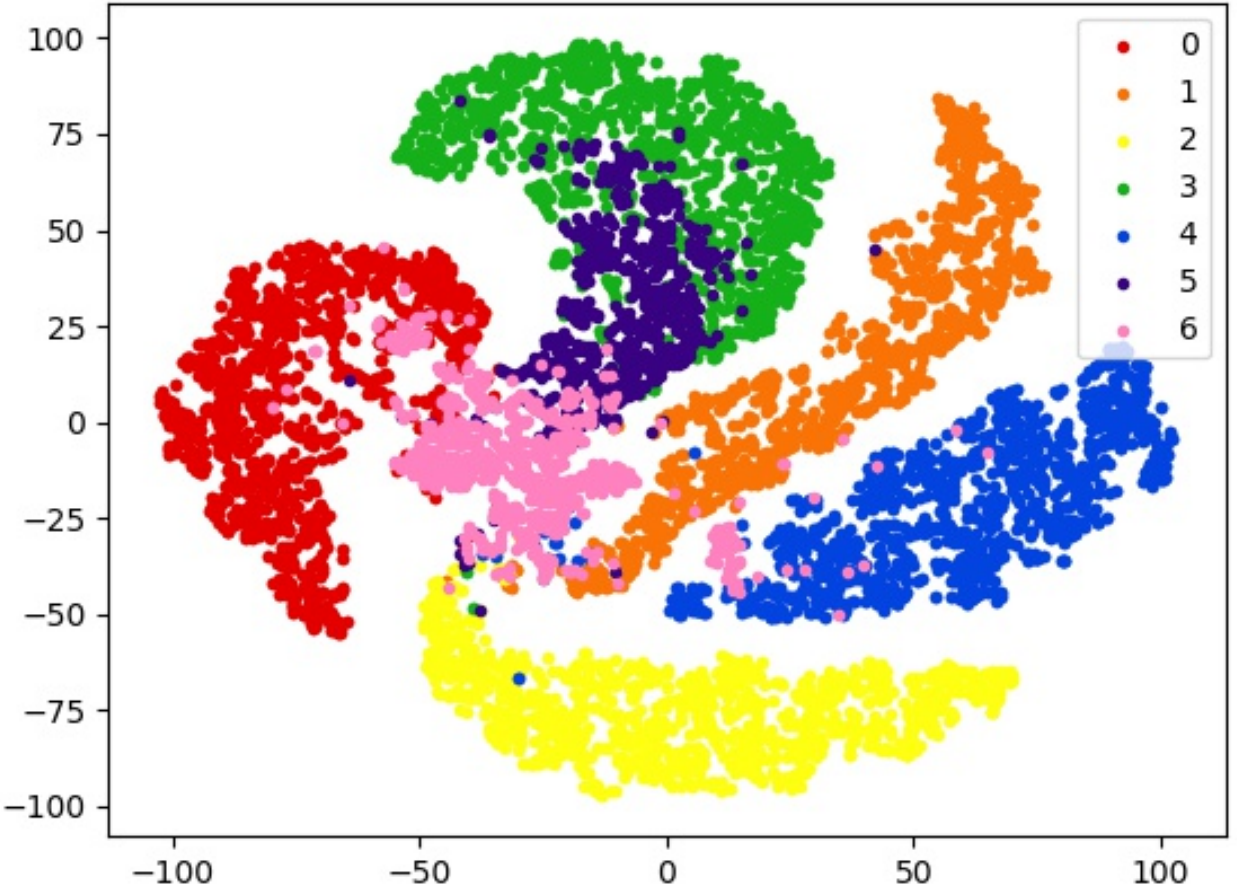}\\
			\mbox{  ({\it a}) {DM on MNIST}}
		\end{minipage}
		\begin{minipage}[h]{33mm}
			\centering
			\includegraphics[width=33mm]{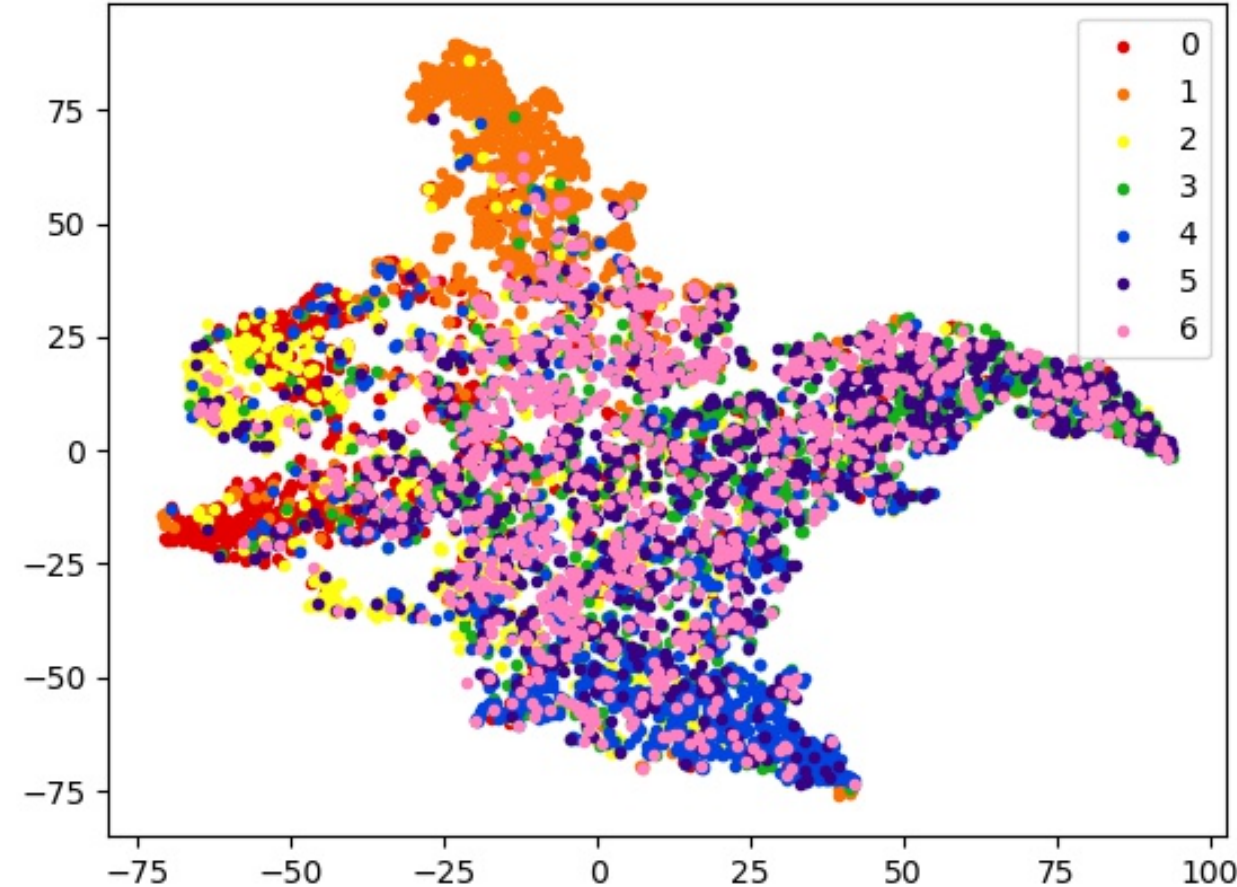}\\
			\mbox{  ({\it b}) {DM on CIFAR-10}}
		\end{minipage}
		\begin{minipage}[h]{33mm}
			\centering
			\includegraphics[width=33mm]{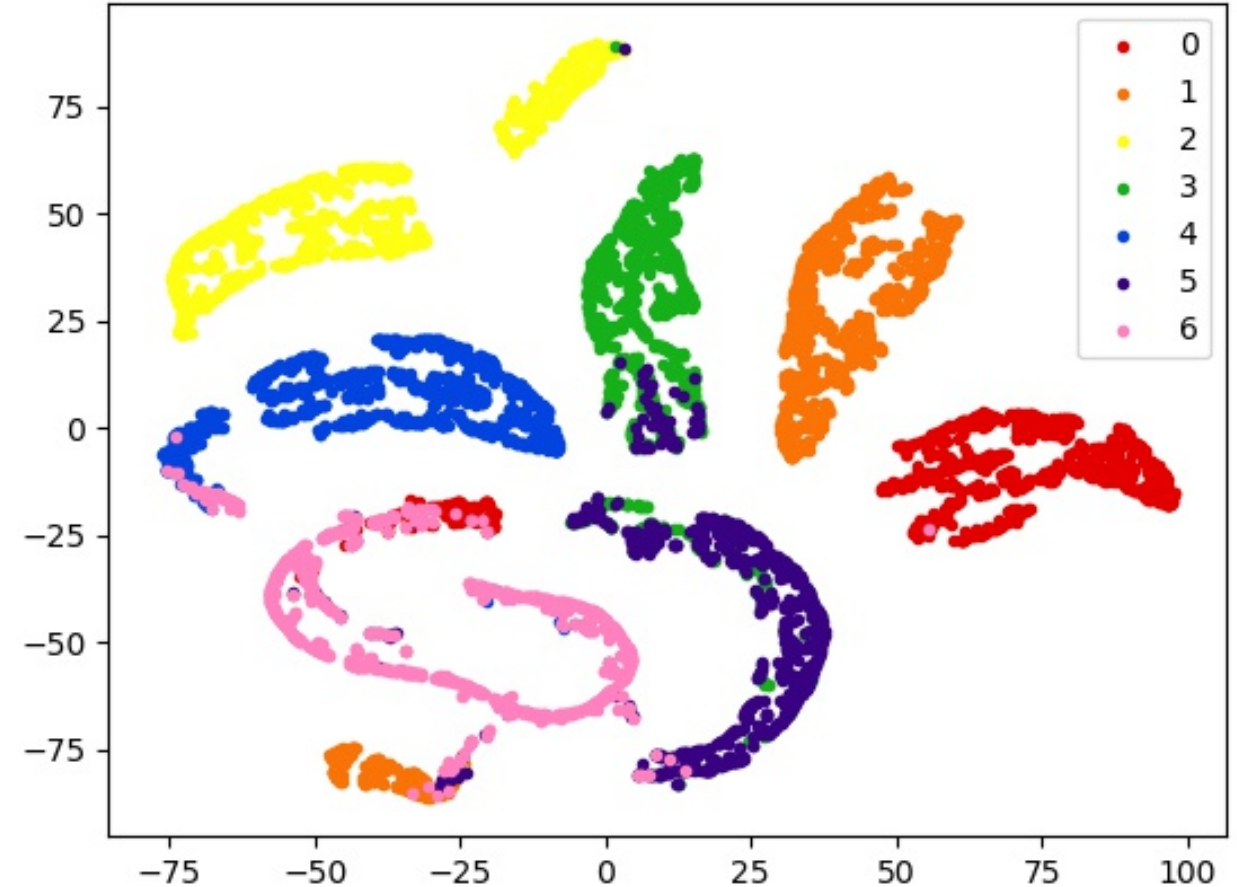}\\
			\mbox{ ({\it c}) {GM on MNIST}}
		\end{minipage}
		\begin{minipage}[h]{33mm}
			\centering
			\includegraphics[width=33mm]{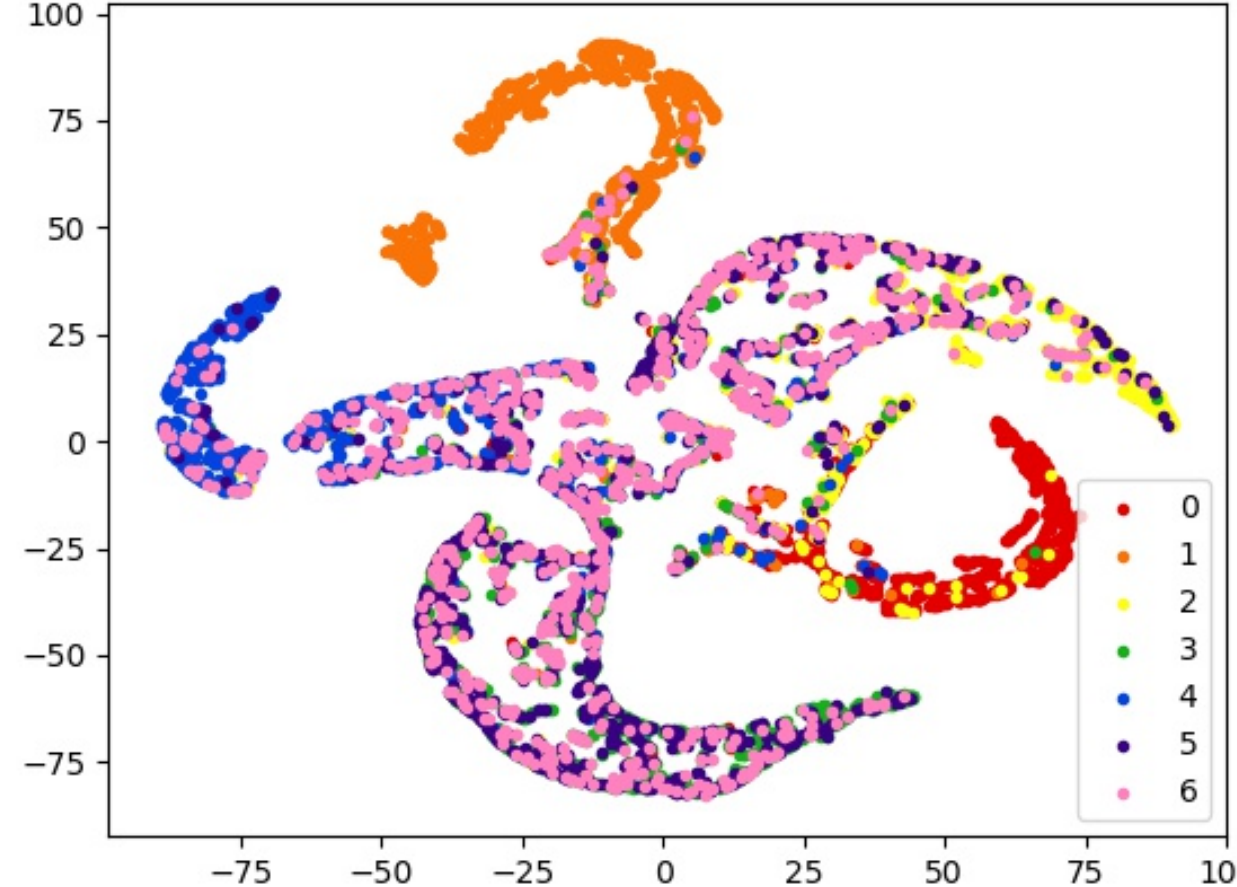}\\
			\mbox{ ({\it d}) {GM on CIFAR-10}}
		\end{minipage}
	\end{center}
	\caption{T-SNE of discriminative model (DM)~\cite{WangKCTK19} and generative model (GM)~\cite{NealOFWL18} on simple (MNIST) and complex (CIFAR-10) datasets. We develop these two models with five classes (i.e., 0-4) in training stage according to the raw paper, then utilize the pre-trained models to achieve the embeddings of known classes (i.e., 0-4) and unknown classes (i.e., 5 and 6) appearing in testing stage. Note that traditional OSC methods usually classify based on the learned embeddings.}\label{fig:f2}
\end{figure}

%, either labeled examples or semantic side-information during training
Therefore, a more realistic classification is to detect out-of-class without any information of unknown classes. With the advent of deep learning, recent OSC approaches can mainly be divided into two aspects: discriminative and generative models. Discriminative models mainly utilize the powerful feature learning and prediction capabilities of deep models to design corresponding distance or prediction confidence measures~\cite{HendrycksG17,WangKCTK19}. In contrast, generative models mainly employ the adversarial learning to generate out-of-class instances near the decision margin that can fool the discriminative model~\cite{GeDG17,JoKKKC18}. In summary, existing OSC approaches focus on learning a representative latent space for in-class examples that preserves details of the given classes. In this case, it is assumed that when presents out-of-class instances to the pre-trained network, it will generate a poor embedding that reports a relatively higher classification error. However, this assumption does not hold for all situations. For example, as shown in Figure \ref{fig:f2}, experiments on MNIST suggest that networks (discriminative and generative models~\cite{WangKCTK19,NealOFWL18}) trained with simple content have high novelty detection accuracy, i.e., the embeddings of out-of-class digits 5 and 6 are well separated from in-class examples. In contrast, instances with complex content, such as CIFAR-10, have much low novelty detection accuracy. This is because latent embeddings learned for in-class examples can also inherently apply to represent some out-of-class instances, for example, the latent embeddings learned for cat (number $3$ in Figure \ref{fig:f2} (b) and (d)) are also able to represent some instances of other out-of-class animal such as dog (number $5$ in Figure \ref{fig:f2} (b) and (d)), considering similar appearance, color and other information. This phenomenon is defined as \textit{Embedding Confusion} in this paper.

We note that out-of-class instances always include confused and distinct ones. Thereby we can firstly select the distinct out-of-class instances, then re-train a new classifier by combing stored in-class examples in semi-supervised paradigm. Meanwhile, the pre-trained model can be employed as a teacher model, which not only ensures that in-class instances are well represented, but also guarantees that out-of-class instances are poorly represented. In result, the learned classifier can obtain well separated embeddings for in-class and out-of-class instances, and significantly improve the detection performance in return. Motivated by this intuition, we propose Semi-Supervised Open Set Classification algorithm (S2OSC), a transductive classifier learning process to mitigate embedding confusion. To the best of our knowledge, none of the previous work has addressed this detection manner. At a high-level, S2OSC can also be adapted to incremental OSC conveniently.

\vspace{-0.5cm}
\section{Related Work}
To set forth the S2OSC, we first introduce existing methods for OSC, which are related to our S2OSC, i.e., discriminative and generative models. Then we present traditional anomaly detection and zero-shot learning methods.

\textbf{Discriminative OSC models.} These approaches mainly restrict intra-class and inter-class distance property on training data, and detect unknown classes by identifying outliers. For example, \cite{DaYZ14} developed a SVM-based method, which learned the concept of known classes while incorporating the structure presented in unlabeled data from open set; \cite{MuZDLZ17} proposed to dynamically maintain two low-dimensional matrix sketches to detect emerging new classes. However, these linear approaches are difficult to process high dimensional space. Recently, several studies have applied deep learning techniques to OSC scenario. For example, \cite{HendrycksG17} distinguished known/unknown class with softmax output probabilities; \cite{LiangLS18} directly utilized temperature scaling to separate the softmax score between in-distribution and out-of-distribution images; \cite{WangKCTK19} proposed a cnn-based prototype ensemble method, which adaptively updates prototype for robust detection. However, these methods can hardly consider the out-of-class instances in training phase.

\textbf{Generative OSC models.} The key component of generation-based OSC models is to generate effective out-of-class instances. For example, \cite{GeDG17} proposed the generative OpenMax (G-OpenMax) algorithm, which provides probability estimation over generated out-of-class instances, that enables the classifier to locate the decision margin according to both in-class and out-of-class knowledge; \cite{JoKKKC18} adopted the GAN technique to generate fake data considering representativeness as the out-of-class data, which can further enhance the robustness of classifier for detection; \cite{NealOFWL18} introduced an augmentation technique, which adopts an encoder-decoder GAN architecture to generate synthetic instances similar to known classes. Though these methods have achieved some improvements, generating more effective out-of-class instances with complex content still need further research~\cite{NealOFWL18}.
 
\textbf{Traditional detection models.} Anomaly detection and generalized zero-shot learning are also related to OSC task. The goal of anomaly detection is to separate outlier instances, for example, \cite{LiuTZ08} proposed a non-parametric method IForest, which detects outliers with ensemble trees. However, anomaly detection follows different protocols from OSC methods, and unable to subdivide known classes. Generalized zero-shot learning aims to classify known and unknown classes with side information. For example, \cite{ChangpinyoCGS16} employed manifold learning to align semantic space with visual features; \cite{LiJLZYH19} introduced the feature confusion GAN, which adopts a boundary loss to maximize the margin of known and unknown classes. However, they assume that semantic information of unknown classes is already in existence, which is incomparable with OSC methods.

\section{Our Algorithm: S2OSC}
%In this section, we formalize the problem of open set classification, and give the details of proposed S2OSC, our holistic semi-supervised open set classification method, which incorporates the dominant components of OSC discussed in section~\ref{sec:s2}.
\subsection{Problem Definition}
Without any loss of generality, suppose we have a supervised training set $\D_{tr}=\{(\x_i,\y_i)\}_{i=1}^{N_{tr}}$ at initial time, where $\x_i \in \R^d$ denotes the $i-$th instance, and $\y_i \in Y = \{1,2,\cdots,C\}$ denotes the corresponding label. Then, we receive a pool of unlabeled testing data $\D_{te} = \{(\x_j)\}_{j=1}^{N_{te}}$, where $\x_j \in \R^d$ denotes the $j-$th instance, and label $\y_j \in \hat{Y} =  \{1,2,\cdots,C,C+1,\cdots,C+B\}$ is unknown. $\{1,2,\cdots,C\}$ denotes in-class set and $\{C+1,\cdots,C+B\}$ represents out-of-class set. Therefore, open set classification can be defined as:

\begin{defn}\label{def:d1}
	\textbf{Open Set Classification (OSC)}  With the initial training set $\D_{tr}=\{(\x_i,\y_i)\}_{i=1}^{N_{tr}}$, we aim to construct a model $f: X \rightarrow Y$. Then with the pre-trained model $f$, OSC is to classify the in-class and out-of-class instances in $\D_{te}$ accurately.
\end{defn}

Following most OSC approaches~\cite{DaYZ14,MuZDLZ17,NealOFWL18,LiangLS18,WangKCTK19,geng2020}, we firstly consider all unknown classes as a super-class for detection, then employ unsupervised clustering techniques such as k-means for subdividing (\textit{out-of-class specifically refers to super-class in following}). Therefore, given the $f$ and $\D_{te}$, we transform to build a new classifier $g$ in transductive manner for operating OSC on $\D_{te}$. In detail, S2OSC pre-trains a classification model $f$ with $\D_{tr}$ and stores limited in-class examples $\D_{in}$ from $\D_{tr}$. $f$ is then used for filtering distinct out-of-class instances $\D_{out}$ in $\D_{te}$. After this, we possess in-class and potential out-of-class labeled data $\X = \{\D_{in}, \D_{out}\}$, and unlabeled data $\U = \{\D_{te} \setminus \D_{out}\}$, thereby we can develop a new classifier $g$ in semi-supervised paradigm. Note that there are two ways to train $g$: 1) fine-tuning based on $f$ directly; 2) retraining from scratch while using $f$ as a teacher for knowledge distillation. We select the second way considering the efficiency and effectiveness. Consequently, we acquire the classification results of $\D_{te}$ using learned $g$ in a transductive manner. In fact, S2OSC comprehensively considers the ideas of both discriminant and generative methods, i.e., trying to separate known classes as far as possible, while taking into account the potential information of unknown classes. Next, we describe each part of S2OSC.

%: $\ell = \big[m+d(i,p)-\min\limits_{\x_{n_c} \in \Omega_i}d_{i,n_c}\big]_+$, where $d_{i,j} =  \|f(\x_i)- f(\x_j)\|_2$ is distance measure. $\Omega_i = (\x_i,\x_p,\x_{n_1},\cdots,\x_{n_{C-1}})$ represents hard triplet, in which $\x_p$ is the positive instance of $\x_i$, others are negative instances, $m$ is the margin
\subsection{Data Filtering}
With the initial in-class training data $\D_{tr}$, we firstly develop a deep classification model $f$ as many typical supervised methods:
\begin{equation}\label{eq:e1}
\begin{split}
f^* = \arg\min\limits_{f} \sum_{i=1}^{N_{tr}} \ell(\y_i,f(\x_i))
\end{split}
\end{equation}
$\ell$ can be any convex loss function, and we define as cross-entropy loss for unification here. Meanwhile, we randomly select $K$ examples from each class to constitute $\D_{in}$. Then, we evaluate the weight of each instance in $\D_{te}$ by self-taught weighting function. In detail, we compute confidence score for each instance $\x_j$ in $\D_{te}$ using pre-trained model $f^*$: 
\begin{equation}\label{eq:weight}
\begin{split}
w_j = u_j + \lambda d_j
\end{split}
\end{equation}
where $\lambda$ is a fixed hyperparameter. $u_j$ denotes statistic prediction confidence, which is done explicitly with the entropy: $u_j = -\sum_c f_c^*(\x_j)\log f_c^*(\x_j)$. $d_j$ represents statistic distance to each in-class center, i.e., $d_j = \min(\|e_j - \mu_c\|_2^2)$, where $e_j$ represents embeddings extracted from feature output layer of $f^*$, and $\mu_c = \frac{1}{|\D^c_{tr}|} \sum_{\x \in \D^c_{tr}} e_{\x}$ represents $c-$th in-class center, $\D^c_{tr}$ is $c-$th class set. It is notable that highly unsure out-of-class instances have larger weights, while in-class and confused instances have lower weights. In result, we can sort $\D_{te}$ according to $w$, and acquire filtered instances set $\D_{out}$ with the same number $K$ as in-class set, the corresponding super-class is $C'$. Therefore, we have owned in-class and out-of-class labeled data $\X = \{\D_{in}, \D_{out}\}$, unlabeled data $\U = \{\D_{te} \setminus \D_{out}\}$, and aim to develop the new classifier $g$.

%~\cite{SajjadiJT16}
\subsection{Objective Function}
Inspired from~\cite{Kihyuk2020}, we combine two common semi-supervised methods to learn $g$: consistency regularization and pseudo-labeling, which aim to effectively utilize unlabeled data by ensuring the consistency among different data-augmented forms. S2OSC has two contributions: 1) Pseudo-labeling threshold. For a given unlabeled instance, the pseudo-label is only retained if $g$ produces a high-confidence prediction; 2) Pre-trained model teaching. For a given instance, we use the pre-trained model $f$ for knowledge distillation of predictions from known classes. Therefore, we can further separate the confused out-of-class instances with in-class instances.

Specifically, the loss function of $g$ exclusively consists of two terms: a supervised loss $L_s$ applied to labeled data and an unsupervised loss $L_u$. $L_s$ can be represented as:
\begin{equation}\label{eq:e2}
\begin{split}
L_s  = &\frac{1}{2|\X|}\sum_{l=1}^{|\X|} (\ell_{s1}(\x_l,\y_l) + \alpha \ell_{s2}(\x_l,f^*(\x_l)))\\
\ell_{s1} = & H_{in}(\y_l,g(\x_l)) + {\bf 1}_{max(g(\x_l)) \geq \tau} H_{out}(\y_l,g(\x_l))\\
\ell_{s2}  = & KL(f^*(\x_l)\|g_{\setminus C'}(\x_l))
\end{split}
\end{equation}
where $H_{(\cdot)}(p,q) = - p\log q$ is standard cross-entropy loss, $KL(p\|q) = p \log \frac{p}{q}$ denotes KL-divergence. $\alpha$ is a hyperparameter, $\tau$ is a scalar parameter denoting the threshold. $g_{\setminus C'}(\x_l)$ is the prediction distribution with re-softmax except out-of-class $C'$. $\ell_{s1}$ adopts the standard cross-entropy loss, note that there may still have embedding confused known class data in $\D_{out}$, thus we utilize ${\bf 1}_{max(g(\x_l)) \geq \tau}$ term to produce a valid ``one-hot'' probability distribution. Meanwhile, ideally, for labeled known class data in $\X$, $f$ can also produce confident probability distribution, otherwise $f$ tends to predict uniform distribution. Thus, $\ell_{s2}$ receives the soft targets from $f$ for in-class and out-of-class examples, which aim to proceed knowledge distillation by restraining two prediction distribution. $f^*(\x_l)$ and $g_{\setminus C'}(\x_l)$ is with Softmax-T that sharpens distribution by adjusting its temperature $T$ following~\cite{HintonVD15}, i.e., raising all probabilities to a power of $\frac{1}{T}$ and re-normalizing.

%$KL(g(\x_l)\|f^*(\x_l))$ is more suitable for noise handling
%\ell_{u2}  = & {\bf 1}_{\x_u \in \D_{in}} H(f^*(\x_u),g(\Phi(\x_u))) + {\bf 1}_{\x_u \in \D_{out}} KL(f^*(\x_u)\|g(\Phi(\x_u)))
%For unlabeled data, S2OSC firstly obtains the pseudo-label by computing the prediction of weakly augmented version for a given unlabeled instance: $q_u = g(\phi(\x_u))$, $\hat{q}_u = \arg\max(q_u)$ is the pseudo-label. Thus, we enforce the loss against model’s output for a strongly augmented version of $\x_u$:
For unlabeled data, S2OSC firstly obtains the pseudo-label by computing the prediction for a given unlabeled instance: $q_u = g(\x_u)$, and $\hat{q}_u = \arg\max(q_u)$ is the pseudo-label, which is then used to enforce the loss against model’s output for a augmented version of $\x_u$:
\begin{equation}\label{eq:e3}
\begin{split}
L_u  = &\frac{1}{2|\U|}\sum_{u=1}^{|\U|} {\bf 1}_{max(q_u) \geq \tau} (\ell_{u1}(\x_u,\hat{q}_u) + \alpha \ell_{u2}(\x_u,f^*(\x_u)))\\
\ell_{u1} = &  H(\hat{q}_u,g(\Phi(\x_u)))\\
\ell_{u2}  = & KL(f^*(\x_u)\|g_{\setminus C'}(\Phi(\x_u)))
\end{split}
\end{equation}
where $\tau$ denotes threshold above which we retrain the pseudo-label. $\Phi$ represents weak augmentation using a standard flip-and-shift strategy or strong augmentation using CTAugment~\cite{BerthelotCCKSZR20} with Cutout~\cite{Terrance} as~\cite{Kihyuk2020}, we employ the former strategy considering efficiency. See Appendix for details. $\ell_{u2}$ employs similar function on unlabeled data. Thus, $L_u$ encourage the model’s predictions to be low-entropy (i.e., high-confidence) on unlabeled data combing hard-label and soft-label. In summary, the loss minimized by S2OSC is: $L = L_s + \lambda_u L_u$, where $\lambda_u$ is a fixed scalar hyperparameter denoting the relative weight. Consequently, the trained $g$ can classify in-class or out-of-class instances in $D_{te}$, and then employ clustering on out-of-class instances to acquire sub-classes.

\section{Incremental S2OSC (I-S2OSC)}
In this section, we aim to demonstrate that S2OSC can be extended into incremental open set classification scenario conveniently.

\subsection{Problem Definition}
In real applications, we always receive the streaming data, and unknown classes are also emerge incrementally. Thus more generalized setting is incremental OSC, which has two characteristics: 1) Data pool. At time window $t$, we only get the data of current time window, i.e., $\D_{te}^t$, not full amount of previous data; 2) Unknown class continuity. At time window $t$, unknown classes appear partially, thereby we need to incrementally conduct OSC, i.e., every time after receiving the data of time window $t$, OSC is performed. Specifically, the streaming data $\D$ can be divided into $\D = \{\D^t\}_{t=0}^T$, where $\D^0 = \D_{tr}$ is the initial training set. $\D^t = \{\x_j^t\}_{j=1}^{N_t}, t \geq 1$ is with $N_t$ unlabeled instances, and the underlying label $\y_j^t \in \hat{Y}^t$ is unknown, $\hat{Y}^t = \hat{Y}^{t-1} \cup Y^t$, where $\hat{Y}^{t-1}$ is the cumulative known classes until $(t-1)-$th time window and $Y^t$ is the unknown class set in $t-$th window. Therefore, we provide the definition of incremental open set classification:

\begin{defn}\label{def:d2}
	\textbf{Incremental Open Set Classification (IOSC)} At time $t \in \{1,2, \cdots,T\}$, we have pre-trained model $f^{t-1}$ and limited stored in-class examples $M^{t-1}$ until $(t-1)-$th time, then receive newly coming data pool $\D^t$. First, we aim to classify known and unknown classes in $\D^t$ as Definition~\ref{def:d1}. Then, with the labeled data from novel classes and stored data $M^{t-1}$, we update the model while mitigating forgetting to acquire $f^t$. Cycle this process until terminated.
\end{defn}

S2OSC can be applied directly for OSC of $\D^t$ at $t-$th time window, then the extra challenge is to update the model while mitigating forgetting~\cite{ratcliff1990connectionist} of previous in-class knowledge. 

\subsection{Model Update}
There exist two labeling cases after OSC, i.e., manually labeling and self-taught labeling~\cite{MuZDLZ17}. We consider first setting following most approaches~\cite{geng2020,MuZDLZ17,WangKCTK19} to avoid label noise accumulation. In detail, after known/unknown classification operator, we can achieve potential out-of-class instances to query their true labels. However, there exist catastrophic forgetting of known classes if we only use the new data to update the model. To solve this problem, we employ a mechanism to incorporate the stored memory and novel class information incrementally, which can mitigate forgetting of discriminatory characteristics about known classes. Specifically, we utilize the exemplary data $M^{t-1}$ for regularization in fine-tuning:
\begin{equation}\label{eq:update}
\begin{split}
L^t & = \sum_{l} \ell(\y_l,f^t(\x_l))\\
s.t. & \quad \ell(M^{t-1},f^{t}) \leq \ell(M^{t-1},f^{t-1}) 
\end{split}
\end{equation}
The loss term encourages the labeled unknown class examples to fine-tune $f^{t-1}$ for better performance, while constraint term imposes $M^{t-1}$ for less forgetting of old in-class knowledge. We utilize directly joint optimization on $M^{t-1}$ as iCaRL~\cite{RebuffiKSL17} to optimize variant of Eq. \ref{eq:update}. See Appendix for details. After that, we need to update the $M^{t-1}$ to store key points of unknown classes. If $M^{t-1}$ is not full, we can fill selected instances from unknown class directly. Otherwise, we remove equal instances for each known class, and fill instances for each unknown class.

%$\frac{|Y^t||M|}{|\hat{Y}^{t-1}||\hat{Y}^{t}|}$
%$\frac{|M|}{|\hat{Y}^{t}|}$
\begin{table*}[t]{\small
		\centering
		\caption{Comparison of open set classification performance. }
		\label{tab:tab1}
		% \renewcommand\arraystretch{1}
%		\begin{tabular*}{1\textwidth}{@{\extracolsep{\fill}}@{}l|c|c|c|c|c|c|c|c|c|c}
%			\toprule
%			\multirow{2}{*}{Methods} & \multicolumn{5}{c|}{F$_{in}$} & \multicolumn{5}{c}{F$_{out}$}\\
%			\cmidrule(l){2-11}
%			& F-MN & CIFAR & SVHN & MN & CINIC & F-MN & CIFAR & SVHN & MN & CINIC\\
%			\midrule
%			Iforest  &.989 &.861 &.857  &.818 &.824 &.943 &.251  &.230 &.065  &.205\\
%			One-SVM  &.704 &.738 &.638  &.737 &.736 &.467 &.279  &.300 &.461  &.285\\
%			LACU  &.787 &.848 &.879  &.879 &.596 &.398 &.190  &.251 &.223  &.348\\
%			SENC &.906 &.899  &.908 &.899 &.907 &.697  &.110 &.007  &.005  &.005\\
%			\midrule
%			ODIN &.954 &\bf.909 &.953  &.954 &.909 &.391 &.031  &.219 &.390  &-\\
%			CFO &.905 &.905 &.849  &.843 &.849 &.275 &.275  &.285 &.278  &.285\\
%			CPE &.871 &.825 &.841  &.988 &.814 &.382 &.383  &.563 &.923  &.239 \\
%			\midrule
%%			MCL  &.920 &\bf.920 &.880  &.970 &881 &- &-  &- &-  &.052\\
%			DTC &.908 &.805 &.935  &.858 &\bf.984 &.212 &.252  &.142 &.249  &.353\\
%			\midrule
%			S2OSC  &\bf.999 & .882 &\bf.938  &\bf.988 &.848 &\bf.993 &\bf.853  &\bf.932 &\bf.987  &\bf.827\\
%		\end{tabular*}
		\begin{tabular*}{1\textwidth}{@{\extracolsep{\fill}}@{}l|c|c|c|c|c|c|c|c|c|c}
			\toprule
			\multirow{2}{*}{Methods} & \multicolumn{5}{c|}{Accuracy } & \multicolumn{5}{c}{F1 }\\
			\cmidrule(l){2-11}
			& F-MN & CIFAR & SVHN & MN & CINIC & F-MN & CIFAR & SVHN & MN & CINIC\\
			\midrule
			Iforest  &.554 &.243 &.198  &.632 &.240 &.553 &.243  &.197 &.625  &.240\\
			One-SVM  &.474 &.260 &.195  &.537 &.243 &.671 &.223  &.102 &.520  &.206\\
			LACU  &.394 &.325 &.193  &.695 &.303 &.409 &.326  &.091 &.681  &.268\\
			SENC &.420 &.215 &.184  &.358 &.230 &.489 &.171  &.124 &.302  &.166\\
			\midrule
			ODIN &.563 &.426 &.601  &.778 &.329 &.854 &.380  &.584 &.767  &.243\\
			CFO &.639 &.502 &.663  &.514 &.389 &.720 &.514  &.656 &.513  &.359\\
			CPE &.628 &.438 &.645  &.961 &.246 &.605 &.353  &.791 &.960  &.270 \\
			\midrule
%			MCL &.744 &.399 &.469  &.770 &.308 &.704 &.359  &.434 &.681  &.253\\
			DTC &.576 &.363 &.534  &.741 &.388 &.665 &.495  &.606 &.717  &.452 \\
			\midrule
			S2OSC &\bf.972 &\bf.847 &\bf.898  &\bf.985 &\bf.785 &\bf.972 &\bf.854  &\bf.901 &\bf.985  &\bf.787  \\
	\end{tabular*}
		\begin{tabular*}{1\textwidth}{@{\extracolsep{\fill}}@{}l|c|c|c|c|c|c|c|c|c|c}
		\toprule
		\multirow{2}{*}{Methods} & \multicolumn{5}{c|}{Precision} & \multicolumn{5}{c}{Recall}\\
		\cmidrule(l){2-11}
		& F-MN & CIFAR & SVHN & MN & CINIC & F-MN & CIFAR & SVHN & MN & CINIC\\
		\midrule
		Iforest  & .553 & .554 & .252 & .243 & .198 & .197 & .657 & .632 & .245 & .240 \\
		One-SVM  & .671 & .474 & .286 & .260 & .195 & .102 & .616 & .537 & .274 & .243 \\
		LACU  & .409 & .394 & .331 & .325 & .193 & .091 & .676 & .695 & .363 & .303 \\
		SENC & .489 & .420 & .253 & .215 & .184 & .124 & .448 & .358 & .211 & .230 \\
		\midrule
		ODIN & .854 & .563 & .520 & .426 & .601 & .584 & .878 & .778 & .554 & .329 \\
		CFO & .720 & .639 & .579 & .502 & .663 & .656 & .598 & .514 & .436 & .389 \\
		CPE & .605 & .698 & .336 & .408 & .645 & .791 & .955 & .961 & .302 & .316 \\
		\midrule
		DTC &  .665 & .576 & .435 & .463 & .434 & .606 & .699 & .681 & .428 & .388 \\
		\midrule
		S2OSC &  \bf.972 & \bf.972 & \bf.888 & \bf.847 & \bf.898 & \bf .901 & \bf.986 &\bf .985 &\bf .799 & \bf.785 \\
		\bottomrule
	\end{tabular*}}
\end{table*}

\vspace{-0.2cm}
\section{Experiments}
%We validate the effectiveness of S2OSC and I-S2OSC on common open set classification benchmarks (section \ref{sec:s3} and section \ref{sec:s4}). Our ablation study test the contribution of each components (section \ref{sec:s5}).

\subsection{Datasets and Baselines}
Considering that incremental OSC is an extension of OSC, the incremental OSC methods can also be applied to the setting of OSC. Therefore, we adopt commonly used incremental OSC datasets for validation here. In detail, we utilize five visual datasets in this paper following~\cite{WangKCTK19}, i.e., FASHION-MNIST (F-MN)~\cite{Xiao2017}, CIFAR-10 (CIFAR)~\cite{krizhevsky2009learning}, SVHN~\cite{netzer2011reading}, MNIST (MN)~\cite{lecun1998mnist}, CINIC~\footnote{https://github.com/BayesWatch/cinic-10}. To validate the effectiveness of proposed approach, we compared it with existing state-of-the-art OSC and incremental OSC methods. First, we compare it with traditional anomaly detection and linear methods: Iforest~\cite{LiuTZ08}, One-Class SVM (One-SVM)~\cite{ScholkopfPSSW01}, LACU-SVM (LACU)~\cite{DaYZ14}, SENC-MAS (SENC)~\cite{MuZDLZ17}. Second, we compare it with recent deep methods: ODIN-CNN (ODIN)~\cite{LiangLS18}, CFO~\cite{NealOFWL18}, CPE~\cite{WangKCTK19} and DTC~\cite{HanVZ19}. Abbreviations in parentheses. DTC is clustering based methods for multiple unknown classes detection. Note that Iforest, One-SVM, LACU, ODIN, CFO, and DTC are OSC methods, SENC and CPE are incremental OSC methods. All OSC baselines except Iforest can be updated incrementally using newly labeled unknown class data and memory data. 

% MCL~\cite{HsuLSOK19}

%\begin{figure}[t]
%	\begin{center}
%		\begin{minipage}[h]{33mm}
%			\centering
%			\includegraphics[width=33mm]{CPE_MNIST.pdf}\\
%			\mbox{  ({\it a}) {Original}}
%		\end{minipage}
%		\begin{minipage}[h]{33mm}
%			\centering
%			\includegraphics[width=33mm]{CPE_CIFAR10.pdf}\\
%			\mbox{  ({\it b}) {S2OSC}}
%		\end{minipage}
%	\end{center}
%	\caption{T-SNE Visualization for both known and unknown classes on CIFAR10 dataset. (a) original feature space; (b) Learned representations through proposed S2OSC.}\label{fig:f3}
%\end{figure}

\subsection{Open Set Classification}\label{sec:s3}
To rearrange each dataset for emulating a OSC form, we randomly hold out 50$\%$ classes as initial training set, and leave one class for testing. Experiments about OSC with various number of out-of-classes can refer to supplementary materials. Moreover, we extracted 33$\%$ of the known class data into test set, so that the test set mixes with known and unknown classes. Here we utilize four commonly used criteria, i.e., Accuracy, Precision, Recall and F1 (Weighted F1), to measure the classification performance, which considers all known and unknown classes. For example, accuracy $A = \frac{\sum_{i=1}^{|\hat{Y}|}(TP_i+TN_i)}{\sum_{i=1}^{|\hat{Y}|}TP_i+TN_i+FP_i+FN_i}$, where $TP, FP, FN, TN$ denotes the true positives, false positive, false negatives and true negatives.

%For further measuring the discrimination of known and unknown classes, we utilize another 2 criteria as~\cite{WangKCTK19}, which treats OSC as a binary classification problem. In detail, we consider all known classes as positive class and all unknown classes as negative class. $F_{in} = \frac{2TP}{2TP+FP+FN}$ is F1 of known classes, $TP, FP, FN, TN$ denotes the true positives, false positive, false negatives and true negatives. Similarly, we can also get the F1 of unknown classes as $F_{out}$. 

\begin{wrapfigure}{l}{7.5cm}%靠文字内容的左侧
	\begin{center}
		\begin{minipage}[h]{33mm}
			\centering
			\includegraphics[width=33mm]{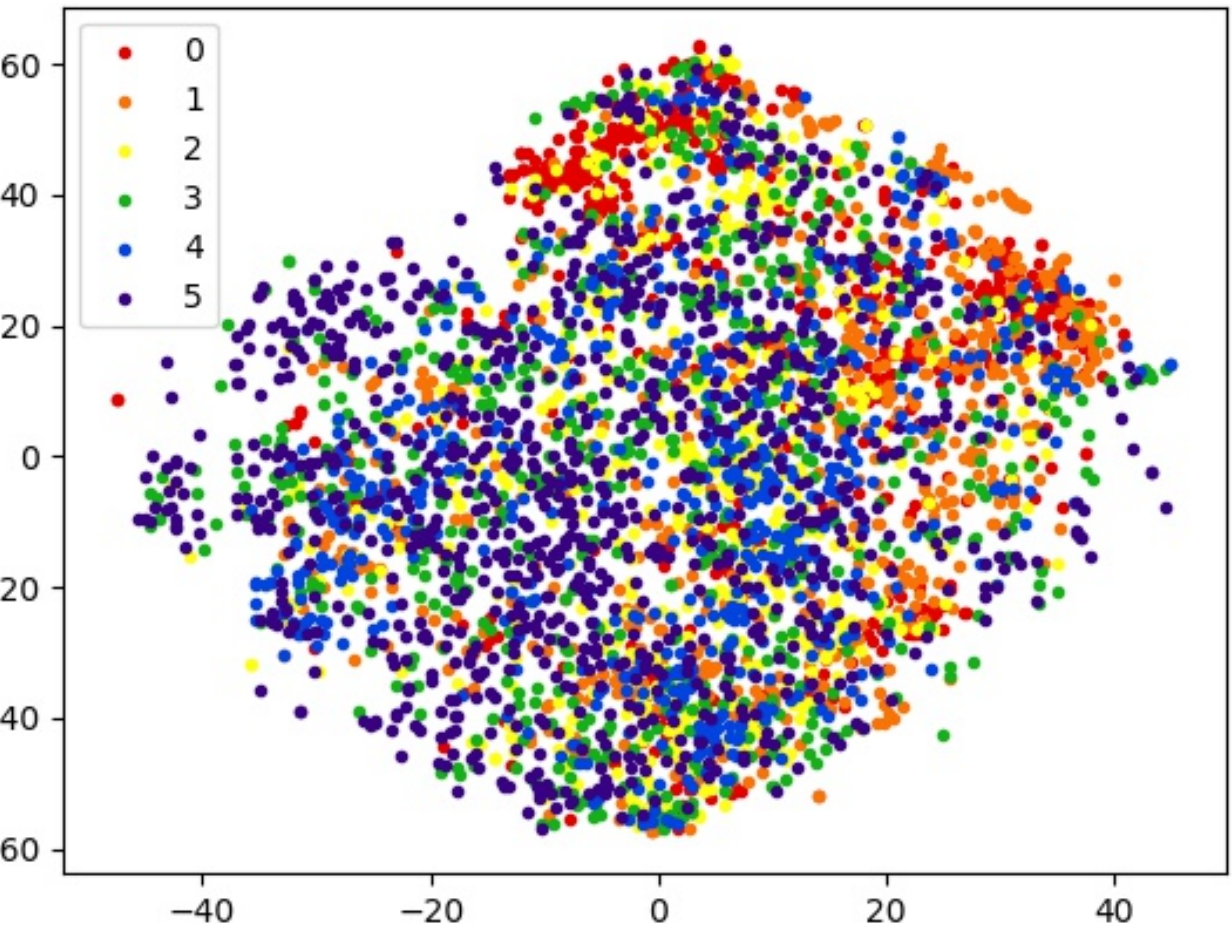}\\
			\mbox{  ({\it a}) {Original}}
		\end{minipage}
		\begin{minipage}[h]{33mm}
			\centering
			\includegraphics[width=33mm]{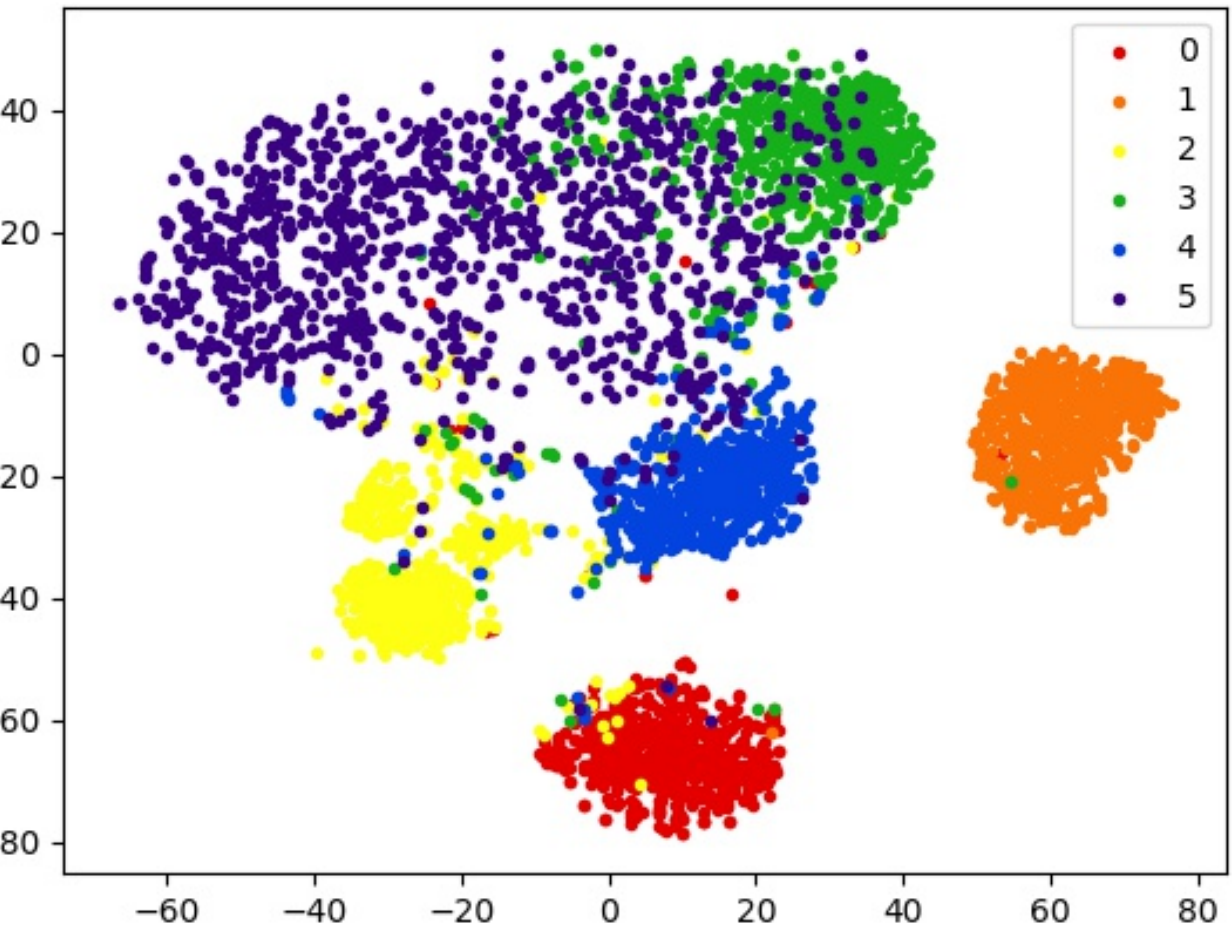}\\
			\mbox{  ({\it b}) {S2OSC}}
		\end{minipage}
	\end{center}
	\caption{T-SNE Visualization on CIFAR-10 dataset. (a) original feature space; (b) learned embeddings by proposed S2OSC.}\label{fig:f3}
\end{wrapfigure}

%``-'' denotes no results
Table \ref{tab:tab1} compares the classification performances of S2OSC with all baselines. We observe that: 1) outlier detection and linear methods perform poorly on most complex datasets, i.e., CIFAR-10, SVHN and CINIC, this indicates that they are difficult to process high dimensional data with complex content; 2) CNN-based methods are better than traditional OSC approaches, i.e., One-SVM, LACU, SENC. This indicates that neural network can provide better feature embeddings for prediction; 3) S2OSC consistently outperforms all baselines over various criteria by a significant margin. For example, in all datasets, S2OSC provides at least 20$\%$ improvements than baselines. This indicates the effectiveness of semi-supervised operation for mitigating embedding confusion. Figure \ref{fig:f3} shows feature embedding results using T-SNE with the similar setting in Figure~\ref{fig:f2}. Clearly, the figure (b) shows that the output of S2OSC has learned distinct groups, which is much better than original embeddings and corresponding embeddings of other deep methods in Figure~\ref{fig:f2}. This validates that instances from unknown classes are well separated from other known clusters, which benefits for unknown class detection. 

%4) The F$_in$ indicator of S2OSC is lower than other methods on some datasets, i.e., CIFAR-10 and CINIC. This is because other methods tend to classify the known classes well, while ignore unknown class detection, i.e., the corresponding F$_{out}$ is poor, the overall S2OSC is still better.

\begin{figure}[htb]
	\begin{center}
		\begin{minipage}[h]{33mm}
			\centering
			\includegraphics[width=33mm]{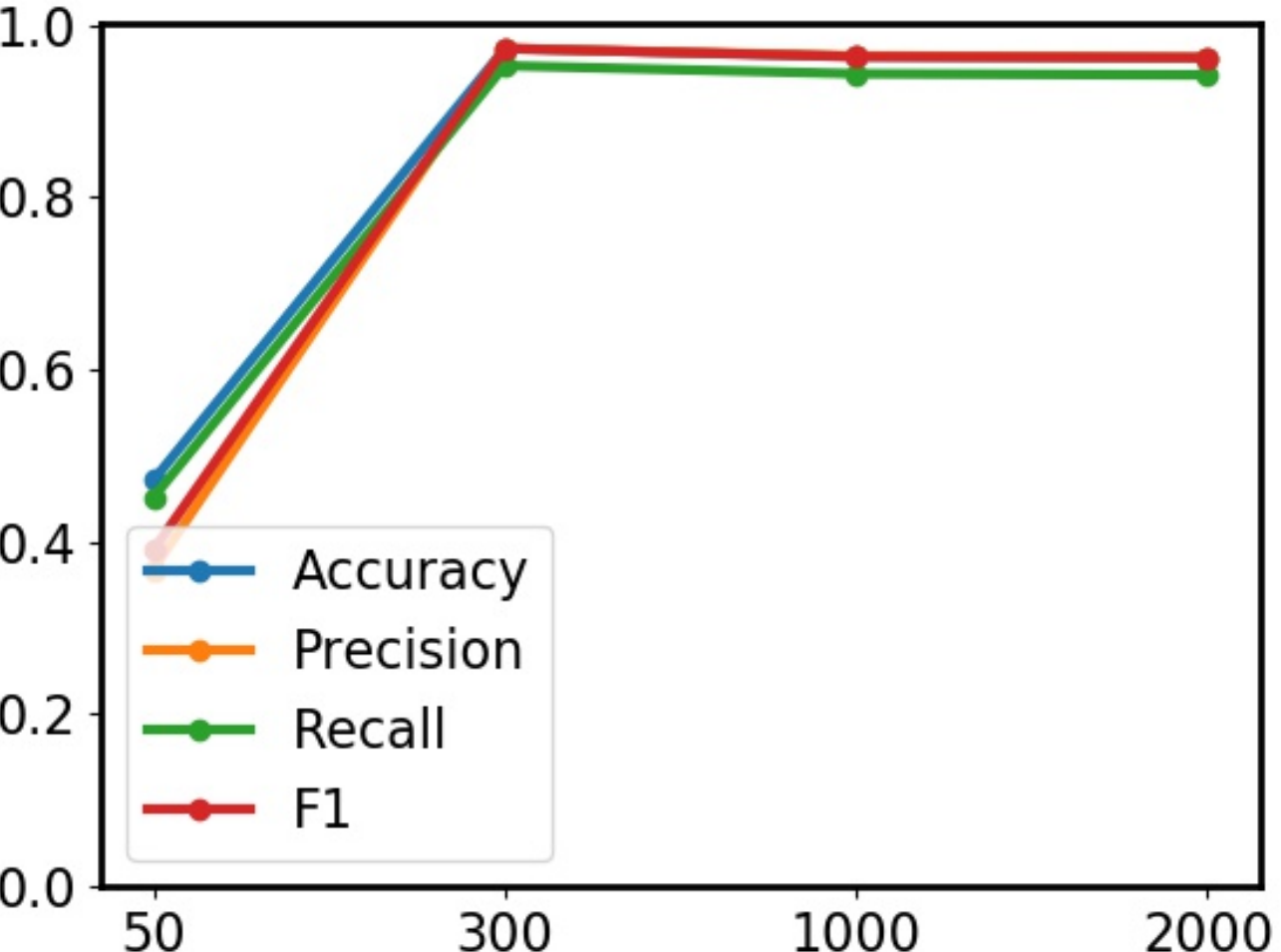}\\
			\mbox{  ({\it a}) {Fashion-MNIST}}
		\end{minipage}
		\begin{minipage}[h]{33mm}
			\centering
			\includegraphics[width=33mm]{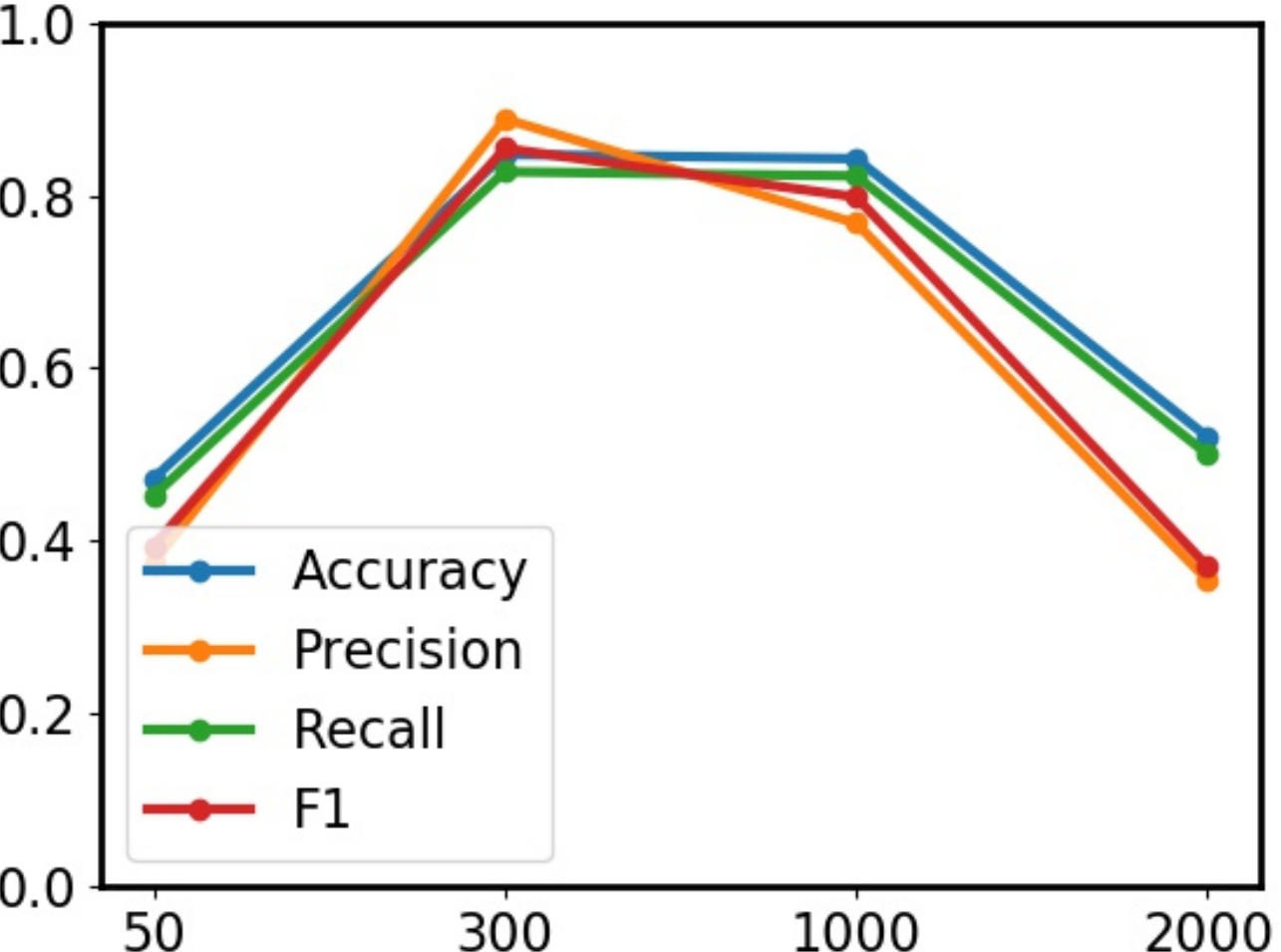}\\
			\mbox{  ({\it b}) {CIFAR-10}}
		\end{minipage}
		\begin{minipage}[h]{33mm}
			\centering
			\includegraphics[width=33mm]{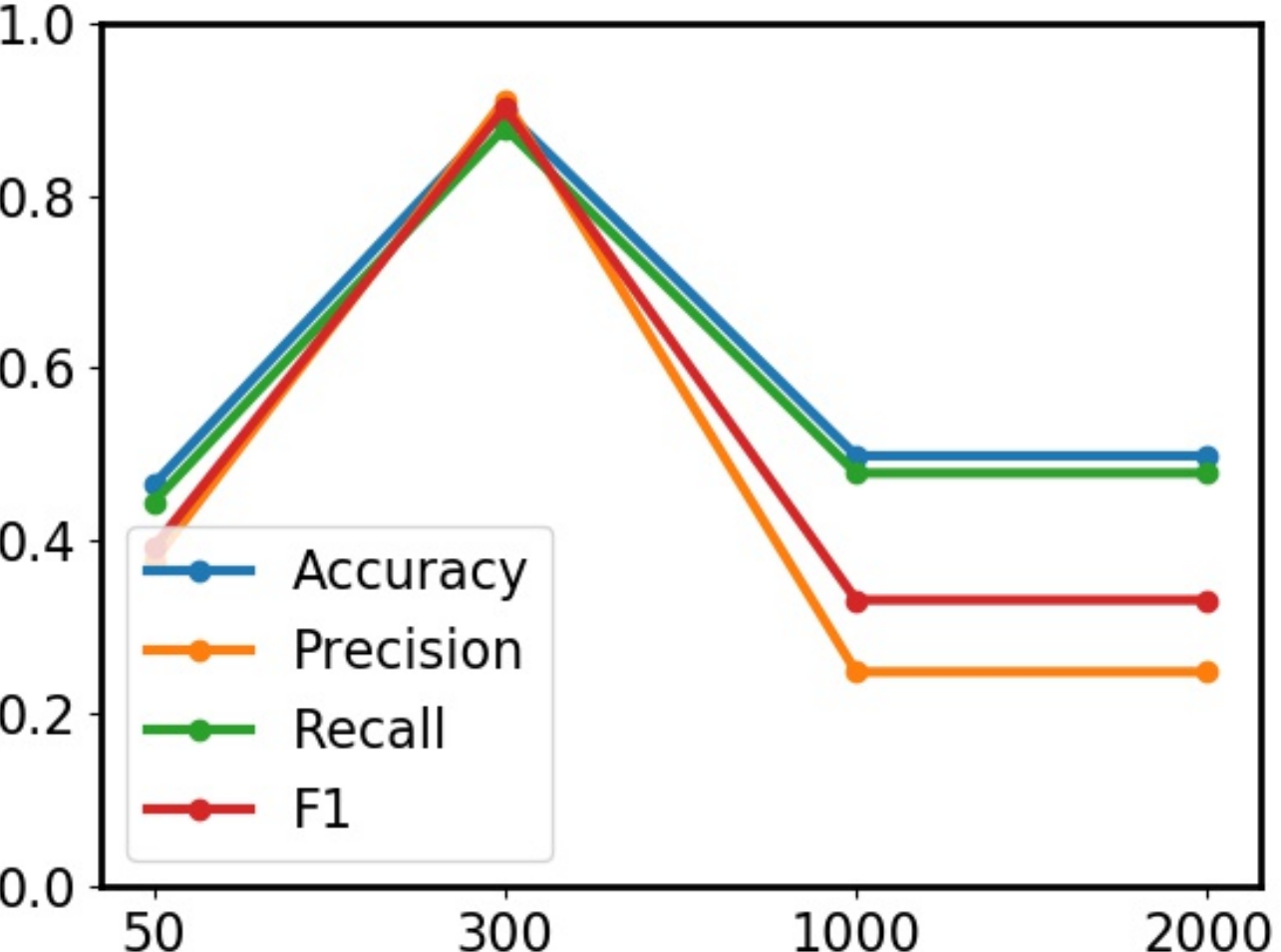}\\
			\mbox{ ({\it c}) {SVNH}}
		\end{minipage}
		\begin{minipage}[h]{33mm}
			\centering
			\includegraphics[width=33mm]{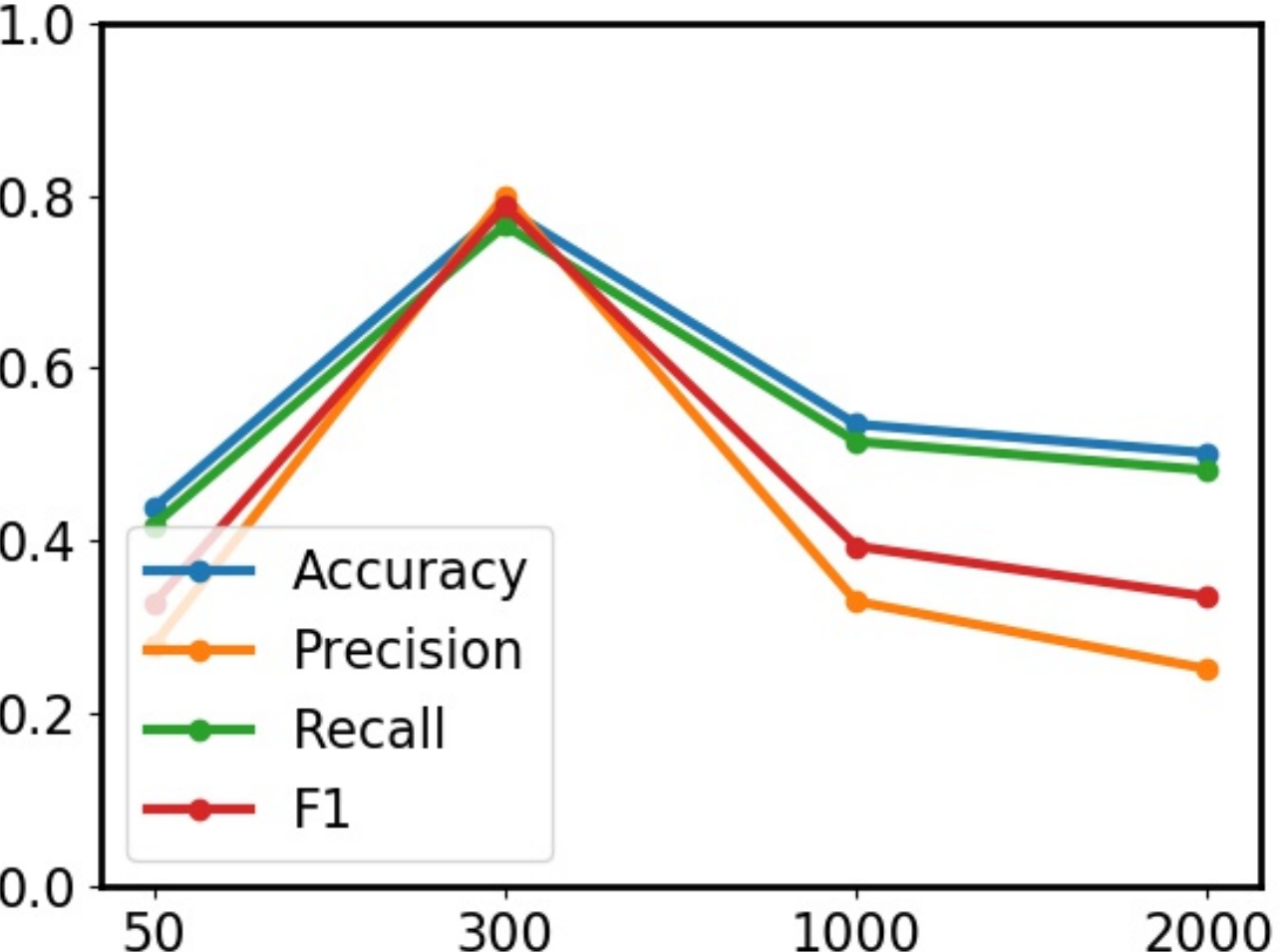}\\
			\mbox{ ({\it d}) {CINIC}}
		\end{minipage}
	\end{center}
	\caption{Classification performance with various number of filtered out-of-class instances.}\label{fig:f1}
\end{figure}

%\subsection{Influence of Filter Size}\label{sec:s5}
Figure \ref{fig:f1} shows the influence of important parameter $K$ (filtering size), i.e., we tune the size of $K = \{50, 300, 1000, 2000\}$. The results reveal that, at first, different criteria improve with the increase of filtered instances. After filtering size is more than a threshold, the performance start to decrease. For example, on CIFAR-10, the accuracy on 300 filtering is about 84.7$\%$, yet the performance decreases after 300 filtering, this is due to the introduction of embedding confused instances with the increase of $K$.

\begin{table*}[t]{\footnotesize
		\centering
		\caption{Comparison of incremental open set classification performance. }
		\label{tab:tab3}
		% \renewcommand\arraystretch{1}
%		\begin{tabular*}{1\textwidth}{@{\extracolsep{\fill}}@{}l|c|c|c|c|c|c|c|c|c|c}
%			\toprule
%			\multirow{2}{*}{Methods} & \multicolumn{5}{c|}{Average F$_{in}$} & \multicolumn{5}{c}{Average F$_{out}$}\\
%			\cmidrule(l){2-11}
%			& F-MN & CIFAR & SVHN & MN & CINIC & F-MN & CIFAR & SVHN & MN & CINIC\\
%			\midrule
%			Iforest  &.948 &.901 &.865  &.919 &.885 &.592 &.260  &.160 &.390  &.123\\
%			One-SVM  &.711 &.876 &.818  &.717 &.852 &.315 &.179  &.111 &.337  &.162\\
%			LACU  &.924 &.918 &.924  &.789 &.868 &.044 &.038  &.050 &.062  &.069\\
%			SENC &.913 &.830 &.899  &.922 &.918 &.312 &.182  &.076 &.062  &.063\\
%			\midrule
%			ODIN &.963 &.931 &.942  &.970 &.930 &.622 &.024  &.286 &.713  &.135\\
%			CFO &.712 &.220 &.437  &.673 &.219 &.411 &.131  &.274 &.453  &.139\\
%			CPE &.931 &.775 &.921  &.988 &.774 &.462 &.373  &.623 &.933  &.249 \\
%			\midrule
%%			MCL  &.930 &.930 &.931  &.928 &.921 & - & -  & - & -  &.052 \\
%			DTC &.918 &.905 &.905  &.928 &.904 &.292 &.302  &.232 &.269  &.263 \\
%			\midrule
%			I-S2OSC &.792 &.793 &.794  &.957 &.753 &.199 &.364  &.372  &.789  &\bf .338 \\
%		\end{tabular*}
		\begin{tabular*}{1\textwidth}{@{\extracolsep{\fill}}@{}l|c|c|c|c|c|c|c|c|c|c}
		\toprule
		\multirow{2}{*}{Methods} & \multicolumn{5}{c|}{Average Accuracy} & \multicolumn{5}{c}{Average F1}\\
		\cmidrule(l){2-11}
		& F-MN & CIFAR & SVHN & MN & CINIC & F-MN & CIFAR & SVHN & MN & CINIC\\
		\midrule
		Iforest  &.533 &.221 &.170  &.606 &.220 &.489 &.217  &.162 &.595  &.217\\
		One-SVM  &.404 &.218 &.167  &.471 &.205 &.409 &.169  &.085 &.476  &.156\\
		LACU  &.332 &.199 &.149  &.171 &.188 &.267 &.137  &.052 &.088  &.114\\
		SENC &.341 &.197 &.156  &.296 &.197 &.264 &.145  &.100 &.251  &.138\\
		\midrule
		ODIN &.780 &.334 &.625  &.853 &.276 &.767 &.284  &.593 &.850  &.206\\
		CFO &.659 &.306 &.485  &.745 &.291 &.635 &.304  &.472 &.722  &.283\\
		CPE &.618 &.368 &.695  &\bf.961 &.286 &.599 &.343  &.701 &\bf.960  &.260\\
		\midrule
%		MCL &.774 &.429 &.509  &.760 &.358 &.718 &.369  &.414 &.681  &.293\\
		DTC &.586 &.393 &.514  &.711 &.348 &.609 &.445  &.566 &.717  &.382 \\
		\midrule
		I-S2OSC &\bf .818 &\bf.660 &.\bf 771  &.926 &\bf.567 &\bf .774 &\bf.609  &\bf.732  &.913  &\bf.517 \\
	\end{tabular*}
\begin{tabular*}{1\textwidth}{@{\extracolsep{\fill}}@{}l|c|c|c|c|c|c|c|c|c|c}
	\toprule
	\multirow{2}{*}{Methods} & \multicolumn{5}{c|}{Average Precision} & \multicolumn{5}{c}{Average Recall}\\
	\cmidrule(l){2-11}
	& F-MN & CIFAR & SVHN & MN & CINIC & F-MN & CIFAR & SVHN & MN & CINIC\\
	\midrule
	Iforest  &.498 & .503 & .226 & .221 & .166 & .170 & .621 & .607 & .225 & .221\\
	One-SVM  &.597 & .405 & .216 & .219 & .187 & .167 & .672 & .472 & .210 & .205\\
	LACU  &.262 & .332 & .133 & .200 & .043 & .150 & .266 & .172 & .143 & .188\\
	SENC &.330 & .342 & .207 & .198 & .210 & .157 & .390 & .297 & .197 & .198\\
	\midrule
	ODIN & \bf.858 & .781 & .456 & .334 & .672 & .626 & .920 & .853 & .372 & .276\\
	CFO &.634 & .659 & .307 & .306 & .477 & .486 & .803 & .745 & .285 & .291 \\
	CPE &.666 & .619 & .427 & .368 & \bf .762 & .696 &  \bf.965 &  \bf.961 & .332 & .286 \\
	\midrule
	DTC & .666 & .586 & .536 & .394 & .657 & .515 & .800 & .711 & .469 & .349\\
	\midrule
	I-S2OSC &.759 &  \bf.818 &  \bf.597 &  \bf.661 & .718 &  \bf.771 & .893 & .922 &  \bf.509 &  \bf.568\\
	\bottomrule
\end{tabular*}}
\end{table*}

\begin{table}[htb]{\footnotesize
		\centering
		\caption{Forgetting measure of known classes over streaming data.}
		\label{tab:tab2}
		\begin{tabular*}{1\textwidth}{@{\extracolsep{\fill}}@{}l|c|c|c|c|c|c|c|c|c}
			\toprule
			\multirow{2}{*}{Methods} & \multicolumn{9}{c}{Forgetting} \\
			\cmidrule(l){2-10}
			& Iforest & One-SVM & LACU & SENC & ODIN & CFO & CPE & DTC & I-S2OSC\\
			\midrule
			F-MN   & N/A &.139 &.121  &.127 &.107 &.047 &.032  &.055  &\bf .029\\
			CIFAR  & N/A &.202 &.127  &.172 &.132 &.128 &.118  &.120  &\bf .117\\
			SVHN   & N/A &.243 &.330  &.249 &.168 &.130 &.124   &.159  &\bf.123\\
			MN     & N/A &.141 &.080  &.061 &.049 &.040 &\bf.033   &.044  &.037\\
			CINIC  & N/A &.224 &.227  &.278 &.170 &.190 &.147  &.175  &\bf .138\\
			\bottomrule
	\end{tabular*}}
\end{table}

%\begin{table}[htb]{\footnotesize
%		\centering
%		\caption{Forgetting measure of known classes over streaming data.}
%		\label{tab:tab2}
%		% \renewcommand\arraystretch{1}
%		\begin{tabular*}{1\textwidth}{@{\extracolsep{\fill}}@{}l|c|c|c|c|c|c|c|c|c|c}
%			\toprule
%			\multirow{2}{*}{Methods} & \multicolumn{10}{c}{Forgetting} \\
%			\cmidrule(l){2-11}
%			& Iforest & One-SVM & LACU & SENC & ODIN & CFO & CPE & MCL & DTC & I-S2OSC\\
%			\midrule
%			F-MN   &.114 &.139 &.121  &.027 &-.107 &.017 &.012  &-.035 &-.025  &.\\
%			CIFAR  &.020 &.202 &.127  &.072 &.102 &.018 &.018  &-.014 &-.010  &.\\
%			SVHN   &.117 &.143 &.230  &.049 &-.028 &.090 &-.024  &-.029 &.059  &.\\
%			MN     &.087 &.021 &.080  &.131 &-.039 &-.030 &.033  &.077 &.014  &.\\
%			CINIC  &.135 &.124 &.127  &.078 &.130 &.190 &.047  &.107 &.075  &.\\
%			\bottomrule
%	\end{tabular*}}
%\end{table}

\subsection{Incremental Open Set Classification}\label{sec:s4}
Furthermore, we rearrange instances in each dataset to emulate a streaming form with incremental unknown classes as~\cite{WangKCTK19}. We utilize the same four criteria, i.e., average Accuracy, average Precision, average Recall and average F1, over various data pools to measure the performance following~\cite{WangKCTK19}, which aims to calculate the overall performance for streaming data. Moreover, to validate the catastrophic forgetting of $f$, we calculate the performance about forgetting profile of different learning algorithms as~\cite{chaudhry2018riemannian}, which defines the difference between maximum knowledge gained of emerging classes on a particular window throughout the learning process and we currently have about it, the lower difference the better.

%Note that we can also get average F$_{in}$ and F$_{new}$ for overall streaming data.

Table \ref{tab:tab3} compares the classification performance of I-S2OSC with all baselines on streaming data. Table \ref{tab:tab2} compares the forgetting performance. ``N/A'' denotes no results considering that Iforest has no update process. We observe that: 1) comparing with results in Table \ref{tab:tab1}, most average classification metrics of deep methods have improved while linear methods decreased, this indicates that deep models can still effectively distinguish known classes for streaming data, which can further benefit OSC; 2) I-S2OSC is superior than other baselines over accuracy and F1 metrics except MNIST dataset, and other two metrics are competitive. But it is not as obvious as the effect in OSC setting. Besides, the average performance of I-S2OSC decreases comparing OSC setting. These phenomenons are because that we uniformly set $K$ to 300, and with the increase of emerging classes, the number of inclusive in-class in filtering data also increases, which will affect the training of $g$. Thus the value of $K$ needs to be tuned carefully; 3) I-S2OSC has the smallest forgetting except MMIST dataset by considering exemplary regularization, which benefits to preserve known class knowledge.

%A negative number indicates that the current task is helpful for future prediction~\cite{chaudhry2018riemannian}.
 
\vspace{-0.4cm}
\section{Conclusion}
Real-word applications always receive the data with unknown classes, thus it is necessary to promote the open set classification. The key challenge in OSC is to overcome the embedding confusion caused by out-of-class instances. To this end, we propose a holistic semi-supervised OSC algorithm, S2OSC. S2OSC incorporated out-of-class instances filtering and semi-supervised model training in a transductive manner, and integrated in-class pre-trained model for teaching. Moreover, S2OSC can be adapted to incremental OSC setting efficiently. Experiments showed the superior performances of S2OSC and I-S2OSC.

\bibliographystyle{nips}\small
\bibliography{acmart}

\begin{thebibliography}{10}

\bibitem{geng2020}
Geng, C., S.~Huang, S.~Chen.
\newblock Recent advances in open set recognition: {A} survey.
\newblock \emph{CoRR}, abs/1811.08581, 2018.

\bibitem{LiuTZ08}
Liu, F.~T., K.~M. Ting, Z.~Zhou.
\newblock Isolation forest.
\newblock In \emph{ICDM}, pages 413--422. 2008.

\bibitem{XiaCWHS15}
Xia, Y., X.~Cao, F.~Wen, et~al.
\newblock Learning discriminative reconstructions for unsupervised outlier
  removal.
\newblock In \emph{ICCV}, pages 1511--1519. 2015.

\bibitem{ChangpinyoCGS16}
Changpinyo, S., W.-L. Chao, B.~Gong, et~al.
\newblock Synthesized classifiers for zero-shot learning.
\newblock In \emph{CVPR}, pages 5327--5336. 2016.

\bibitem{LiJLZYH19}
Li, J., M.~Jing, K.~Lu, et~al.
\newblock Alleviating feature confusion for generative zero-shot learning.
\newblock In \emph{ACMMM}, pages 1587--1595. 2019.

\bibitem{CaiZTM019}
Cai, X., P.~Zhao, K.~Ting, et~al.
\newblock Nearest neighbor ensembles: An effective method for difficult
  problems in streaming classification with emerging new classes.
\newblock In \emph{ICDM}, pages 970--975. 2019.

\bibitem{HendrycksG17}
Hendrycks, D., K.~Gimpel.
\newblock A baseline for detecting misclassified and out-of-distribution
  examples in neural networks.
\newblock In \emph{ICLR}. 2017.

\bibitem{WangKCTK19}
Wang, Z., Z.~Kong, S.~Chandra, et~al.
\newblock Robust high dimensional stream classification with novel class
  detection.
\newblock In \emph{ICDE}, pages 1418--1429. 2019.

\bibitem{GeDG17}
Ge, Z., S.~Demyanov, R.~Garnavi.
\newblock Generative openmax for multi-class open set classification.
\newblock In \emph{BMVC}. 2017.

\bibitem{JoKKKC18}
Jo, I., J.~Kim, H.~Kang, et~al.
\newblock Open set recognition by regularising classifier with fake data
  generated by generative adversarial networks.
\newblock In \emph{ICASSP}, pages 2686--2690. 2018.

\bibitem{NealOFWL18}
Neal, L., M.~L. Olson, X.~Z. Fern, et~al.
\newblock Open set learning with counterfactual images.
\newblock In \emph{ECCV}, pages 620--635. 2018.

\bibitem{DaYZ14}
Da, Q., Y.~Yu, Z.-H. Zhou.
\newblock Learning with augmented class by exploiting unlabeled data.
\newblock In \emph{AAAI}, pages 1760--1766. 2014.

\bibitem{MuZDLZ17}
Mu, X., F.~Zhu, J.~Du, et~al.
\newblock Streaming classification with emerging new class by class matrix
  sketching.
\newblock In \emph{AAAI}, pages 2373--2379. 2017.

\bibitem{LiangLS18}
Liang, S., Y.~Li, R.~Srikant.
\newblock Enhancing the reliability of out-of-distribution image detection in
  neural networks.
\newblock In \emph{ICLR}. 2018.

\bibitem{Kihyuk2020}
Sohn, K., D.~Berthelot, C.~Li, et~al.
\newblock Fixmatch: Simplifying semi-supervised learning with consistency and
  confidence.
\newblock \emph{CoRR}, abs/2001.07685, 2020.

\bibitem{HintonVD15}
Hinton, G.~E., O.~Vinyals, J.~Dean.
\newblock Distilling the knowledge in a neural network.
\newblock \emph{CoRR}, abs/1503.02531, 2015.

\bibitem{BerthelotCCKSZR20}
Berthelot, D., N.~Carlini, E.~D. Cubuk, et~al.
\newblock Remixmatch: Semi-supervised learning with distribution matching and
  augmentation anchoring.
\newblock In \emph{ICLR}. 2020.

\bibitem{Terrance}
Devries, T., G.~W. Taylor.
\newblock Improved regularization of convolutional neural networks with cutout.
\newblock \emph{CoRR}, abs/1708.04552, 2017.

\bibitem{ratcliff1990connectionist}
Ratcliff, R.
\newblock Connectionist models of recognition memory: constraints imposed by
  learning and forgetting functions.
\newblock \emph{Psychol. Review}, 97(2):285, 1990.

\bibitem{RebuffiKSL17}
Rebuffi, S.-A., A.~Kolesnikov, G.~Sperl, et~al.
\newblock icarl: Incremental classifier and representation learning.
\newblock In \emph{CVPR}, pages 5533--5542. 2017.

\bibitem{Xiao2017}
Xiao, H., K.~Rasul, R.~Vollgraf.
\newblock Fashion-mnist: a novel image dataset for benchmarking machine
  learning algorithms.
\newblock \emph{CoRR}, abs/1708.07747, 2017.

\bibitem{krizhevsky2009learning}
Krizhevsky, A., G.~Hinton, et~al.
\newblock Learning multiple layers of features from tiny images.
\newblock 2009.

\bibitem{netzer2011reading}
Netzer, Y., T.~Wang, A.~Coates, et~al.
\newblock Reading digits in natural images with unsupervised feature learning.
\newblock \emph{NeurIPS Workshop}, 2011(2):5, 2011.

\bibitem{lecun1998mnist}
LeCun, Y., C.~Cortes, C.~J. Burges.
\newblock The mnist database of handwritten digits, 1998.
\newblock \emph{URL http://yann. lecun. com/exdb/mnist}, 10:34, 1998.

\bibitem{ScholkopfPSSW01}
Scholkopf, B., J.~C. Platt, J.~Shawe-Taylor, et~al.
\newblock Estimating the support of a high-dimensional distribution.
\newblock \emph{Neural Computation}, 13(7):1443--1471, 2001.

\bibitem{HsuLSOK19}
Hsu, Y., Z.~Lv, J.~Schlosser, et~al.
\newblock Multi-class classification without multi-class labels.
\newblock In \emph{ICLR}. 2019.

\bibitem{HanVZ19}
Han, K., A.~Vedaldi, A.~Zisserman.
\newblock Learning to discover novel visual categories via deep transfer
  clustering.
\newblock In \emph{ICCV}, pages 8400--8408. 2019.

\bibitem{chaudhry2018riemannian}
Chaudhry, A., P.~K. Dokania, T.~Ajanthan, et~al.
\newblock Riemannian walk for incremental learning: Understanding forgetting
  and intransigence.
\newblock \emph{CoRR}, abs/1801.10112, 2018.

\end{thebibliography}


\begin{thebibliography}{10}

\bibitem{Kihyuk2020}
Sohn, K., D.~Berthelot, C.~Li, et~al.
\newblock Fixmatch: Simplifying semi-supervised learning with consistency and
  confidence.
\newblock \emph{CoRR}, abs/2001.07685, 2020.

\bibitem{BerthelotCCKSZR20}
Berthelot, D., N.~Carlini, E.~D. Cubuk, et~al.
\newblock Remixmatch: Semi-supervised learning with distribution matching and
  augmentation anchoring.
\newblock In \emph{ICLR}. 2020.

\bibitem{CubukZMVL19}
Cubuk, E.~D., B.~Zoph, D.~Mane, et~al.
\newblock Autoaugment: Learning augmentation strategies from data.
\newblock In \emph{CVPR}, pages 113--123. 2019.

\bibitem{ratcliff1990connectionist}
Ratcliff, R.
\newblock Connectionist models of recognition memory: constraints imposed by
  learning and forgetting functions.
\newblock \emph{Psychol. Review}, 97(2):285, 1990.

\bibitem{SongT19}
Song, G., X.~Tan.
\newblock Sequential learning for cross-modal retrieval.
\newblock In \emph{ICCV Workshop}, pages 4531--4539. 2019.

\bibitem{RebuffiKSL17}
Rebuffi, S.-A., A.~Kolesnikov, G.~Sperl, et~al.
\newblock icarl: Incremental classifier and representation learning.
\newblock In \emph{CVPR}, pages 5533--5542. 2017.

\bibitem{KirkpatrickPRVD16}
Kirkpatrick, J., R.~Pascanu, N.~C. Rabinowitz, et~al.
\newblock Overcoming catastrophic forgetting in neural networks.
\newblock \emph{CoRR}, abs/1612.00796, 2016.

\bibitem{LeeKJHZ17}
Lee, S.-W., J.-H. Kim, J.~Jun, et~al.
\newblock Overcoming catastrophic forgetting by incremental moment matching.
\newblock In \emph{NIPS}, pages 4655--4665. 2017.

\bibitem{geng2020}
Geng, C., S.~Huang, S.~Chen.
\newblock Recent advances in open set recognition: {A} survey.
\newblock \emph{CoRR}, abs/1811.08581, 2018.

\bibitem{WangKCTK19}
Wang, Z., Z.~Kong, S.~Chandra, et~al.
\newblock Robust high dimensional stream classification with novel class
  detection.
\newblock In \emph{ICDE}, pages 1418--1429. 2019.

\bibitem{HeZRS16}
He, K., X.~Zhang, S.~Ren, et~al.
\newblock Deep residual learning for image recognition.
\newblock In \emph{CVPR}, pages 770--778. 2016.

\bibitem{SutskeverMDH13}
Sutskever, I., J.~Martens, G.~E. Dahl, et~al.
\newblock On the importance of initialization and momentum in deep learning.
\newblock In \emph{ICML}, pages 1139--1147. 2013.

\bibitem{DaYZ14}
Da, Q., Y.~Yu, Z.-H. Zhou.
\newblock Learning with augmented class by exploiting unlabeled data.
\newblock In \emph{AAAI}, pages 1760--1766. 2014.

\bibitem{Kulis13}
Kulis, B.
\newblock Metric learning: {A} survey.
\newblock \emph{Foundations and Trends in Machine Learning}, 5(4):287--364,
  2013.

\bibitem{LiuWYY16}
Liu, W., Y.~Wen, Z.~Yu, et~al.
\newblock Large-margin softmax loss for convolutional neural networks.
\newblock In \emph{ICML}, pages 507--516. 2016.

\bibitem{maaten2008visualizing}
Maaten, L. v.~d., G.~Hinton.
\newblock Visualizing data using t-sne.
\newblock \emph{Journal of machine learning research}, 9(Nov):2579--2605, 2008.

\bibitem{HanVZ19}
Han, K., A.~Vedaldi, A.~Zisserman.
\newblock Learning to discover novel visual categories via deep transfer
  clustering.
\newblock In \emph{ICCV}, pages 8400--8408. 2019.

\bibitem{chaudhry2018riemannian}
Chaudhry, A., P.~K. Dokania, T.~Ajanthan, et~al.
\newblock Riemannian walk for incremental learning: Understanding forgetting
  and intransigence.
\newblock \emph{CoRR}, abs/1801.10112, 2018.

\end{thebibliography}

\end{document}

\maketitle

\section{The Algorithm Pipeline}

\begin{figure*}[htb]\centering
	\centering
	\includegraphics[width = 120mm]{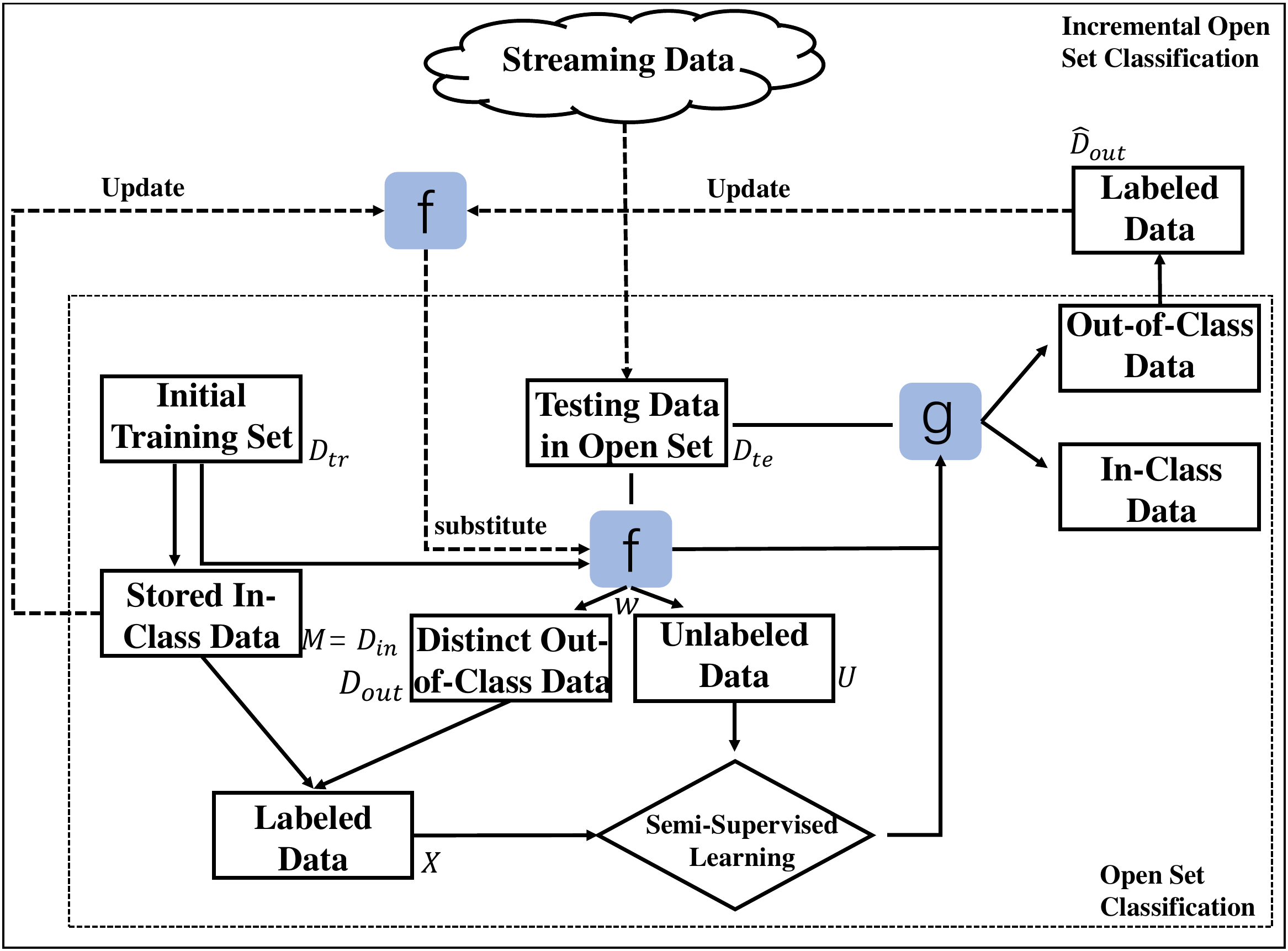}\\
	\caption{Pipeline of the S2OSC and I-S2OSC.}\label{fig:framework}
\end{figure*}

In this section, we provide the pipeline of S2OSC and I-S2OSC as shown in Figure \ref{fig:framework}, where the dotted frame part is the S2OSC pipeline, and the solid line part is the extended I-S2OSC. Specifically, look from top to bottom inside the dotted frame. Given a set of initially in-class training set $\D_{tr}$, we carry on with two jobs: 1) pre-train an in-class classification model $f$; 2) store limited in-class examples  $\D_{in}$. Then, we receive the testing data $\D_{te}$ from open set, and $\D_{te}$ can be divided into two parts using $f$: 1) distinct out-of-class instances $\D_{out}$; 2) unlabeled data $\U = \{\D_{te} \setminus \D_{out}\}$. Here all instances in $\D_{out}$ are reduced to an unified super-class. In result, we possess in-class and out-of-class labeled data $\X = \{\D_{in}, \D_{out}\}$ and unlabeled data $\U = \{\D_{te} \setminus \D_{out}\}$, thereby we can develop a new classifier $g$ in semi-supervised paradigm by considering $f$ as teacher model simultaneously. And we can acquire the classification results of $\D_{te}$ using learned $g$ in a transductive manner. Derived to incremental OSC scenario, we receive the testing data $D^t$ of $t-$th time window from the streaming data, and utilize S2OSC for open set classification. After that, we query the ground-truths of potential unknown class data, and combine stored in-class data to incrementally update $f^t$, then substitute last $f^{t-1}$. It is notable that $g$ is re-trained from scratch for every time window.

\section{Our Algorithm: S2OSC}
\subsection{Weak and Strong Data-augmentation}
In this paper, we leverage the weak and strong augmentation, i.e., $\Phi(\x)$, following~\cite{Kihyuk2020}:
\begin{itemize}
	\item \textbf{Weak augmentation.} It adopts a standard flip-and-shift strategy. On all datasets, Images is randomly flipped horizontally with a probability of 50$\%$, and randomly translated by up to 12.5$\%$ vertically and horizontally;
	\item \textbf{Strong augmentation.} It adopts CTAugment~\cite{BerthelotCCKSZR20} with Cutout technique based on AutoAugment~\cite{CubukZMVL19}. Here, AutoAugment learns an augmentation strategy with reinforcement learning technique, which refers to transformations from the Python Imaging Library\footnote{https://www.pythonware.com/products/pil/}, and requires labeled data. CTAugment is a variant of AutoAugment, which requires no labeled data. Further details on CTAugment can be found in~\cite{BerthelotCCKSZR20}. Cutout is a simple regularization technique that randomly masks out square regions of input image.
\end{itemize}
Therefore, weak augmentation produces slightly distorted version of a given image, while strong augmentation produces heavily distorted version of a given image.

\subsection{S2OSC Considering Multiple Unknown Classes}
As a matter of fact, multiple unknown classes emerging simultaneously have a certain effect on training $g$, that is the pseudo super-class may contains sub-classes with different semantics. Using CIFAR-10 as an example, truck and dog may coexist in unknown class set, and it will lead to separated intra-class distances of unknown super-class. Thus known classes that are semantically similar to unknown classes will affect learning process. To solve this problem, we also turn to unify the known classes into a super-class, which aims to learn the distribution difference between known and unknown class sets. Thus the training of $g$ is transformed into a binary classification problem. In detail, we consider the known class set $\{1,2,3, \cdots, C\}$ as positive class, and unknown class set as negative class. Thus the loss function can be reformulated as:
\begin{equation}\label{eq:e2}
\begin{split}
L = & L_s + \lambda_u L_u \\
L_s  = &\frac{1}{2|\X|}\sum_{l=1}^{|\X|} \{H(\hat{\y}_l,g(\x_l)) + \alpha KL(\psi(\x_l),g(\Phi(\x_l))) \}\\
L_u  = &\frac{1}{2|\U|}\sum_{u=1}^{|\U|} {\bf 1}_{max(q_u) \geq \tau} \{H(\hat{q}_u,g(\Phi(\x_u))) + \alpha KL(\psi(\x_l),g(\Phi(\x_u)))\}\\
\end{split}
\end{equation}
where $\hat{\y}_l \in \{1,0\}$ denotes the re-defined super-class. $H(\cdot)$ denotes the cross-entropy for binary classification. $\psi(\x_l) = [1-\ell(f^*(\x_l)), \ell(f^*(\x_l))]$, $\ell(f^*(\x_l))$ denotes the loss function of $\x$ using $f^*$, $\ell(\cdot)$ is small for known class data, while becomes large for unknown class data. Considering the data balance, a total of $K$ examples are sampled from known classes. With the elementary prediction of learned $g$, we can utilize $f^*$ for further sub-dividing of known classes. In experiment, we also combine labeled data into unlabeled dataset. It is notable that we adopt this S2OSC variant to conduct multi-class OSC experiments, and get superior results.
%1-\ell(f^*(\x_l))

\section{Incremental S2OSC}  
\subsection{Introduction of Forgetting}
In incremental OSC, we need to update model with the labeled instances of novel classes after detecting. Different from re-training with the entire previous in-class data, incremental model update aims to fine-tune the model only referring limited data from known classes. Therefore, the catastrophic forgetting phenomenon becomes an obstacle~\cite{ratcliff1990connectionist}, i.e., we  find that the knowledge learned from previous task (known classes classification) will lose when information relevant to the current task (novel class classification) is incorporated, considering the lack of previous in-class data.

\subsection{Model Update}
{\begin{algorithm}[htb]
		\caption{Model Update}
		\label{alg:alg1}
		\textbf{Input}:\\
		Data: $\hat{\D}_{out}$ //labeled examples of novel classes at time window $t$ \\
		Memory: $M^{t-1}$ // stored examples of known classes at time window $t$ \\
		Model: $f^{t-1}$ // last time model \\
		\begin{algorithmic}[1]{
				\FOR{$\x_j \in M^{t-1}$}
				\STATE $q_j \leftarrow f^{t-1}(\x_j)$ // store network output with pre-trained model
				\ENDFOR
				\WHILE {stop condition is not triggered}
				\FOR{mini-batch}
				\STATE Calculate $L$ according to Equation \ref{eq:e1};
				\STATE Update model parameters of $f^t$ using SGD;
				\ENDFOR
				\ENDWHILE
			}
		\end{algorithmic}
\end{algorithm}}
In this subsection, we will describe incremental model update in detail. After S2OSC operator, we can achieve potential out-of-class instances for querying their true labels. Suppose at time window $t$, we acquire $Q$ examples with ground-truths for novel classes, note that $Q \gg K $, where $K$ is the number of instances in data filtering. And these newly labeled examples constitute $\hat{\D}_{out}$. Therefore, the overall loss for fine-tuning $f$ can be relaxed as: $L = \ell(\hat{\D}_{out},f^t) + \ell(M^{t-1},f^t)$ according to the Eq. 3 in~\cite{SongT19}, which rephrases the constraint term as the task of better performance
on $M^{t-1}$. This can be done with commonly used incremental update approach iCaLR~\cite{RebuffiKSL17}:
\begin{equation}\label{eq:e1}
\begin{split}
L = & -(\sum_{\x_i \in \hat{\D}_{out} \bigcup M^{t-1}} \y_i \log f^t(\x_i) + \sum_{\x_j \in M^{t-1}} \q_j \log f^t(\x_j)) \\
\q_j = & f^{t-1}(\x_j)
\end{split}
\end{equation}
Consequently, the loss function encourages the network to output the correct class indicator (classification loss) for all labeled examples, and reproduces the scores calculated in the previous step (distillation loss) for stored in-class examples. Besides, in the memory update phase, we reduce and add examples according to~\cite{RebuffiKSL17}, which considers the structure representativeness. The details are shown in Algorithm \ref{alg:alg1}.

Here, we adopt the replay-based methods as~\cite{RebuffiKSL17}, rather than use regularization-based methods such as EWC~\cite{KirkpatrickPRVD16}, IMM~\cite{LeeKJHZ17} with extra regularization term on parameters to consolidate previous knowledge. The reason is that regularization-based methods always calculate fisher information matrix~\cite{KirkpatrickPRVD16} for all  parameters in network, which is hard to accomplish for convolutional neural networks. Note that EWC and IMM basically experiment with shallow fully connected network.

\section{Experiments}
%We validate the effectiveness of S2OSC and I-S2OSC on common open set classification benchmarks (section \ref{sec:s3} and section \ref{sec:s4}). Our ablation study test the contribution of each components (section \ref{sec:s5}).
\subsection{Datasets and Baselines}
In this subsection, we will give the details of datasets and baseline methods.

We utilize five publicly visual datasets for evaluating: FASHION-MNIST dataset contains 70,000 images of articles from 10 distinct classes, CIFAR-10 dataset includes 60,000 natural color images of 32x32 pixels from 10 different classes, SVHN dataset also includes 100,000 natural color images of 32x32 pixels about house numbers from 10 different classes, MNIST dataset contains 70,000 labeled handwritten digits images from 10 categories, CINIC scene dataset is enlarged CIFAR-10 with 100,000 images. To validate the effectiveness on dataset with large classes, we further experiment on CIFAR-30 as~\cite{geng2020}, which randomly select 30 classes from CIFAR-100.

For baseline methods, we compare nine state-of-the-art methods including: 1) traditional outlier detection method: Iforest; 2) linear one-class OSC methods: One-Class SVM (One-SVM), LACU-SVM (LACU) and SENC-MAS (SENC); 3) deep one-class OSC methods:  ODIN-CNN (ODIN), CFO and CPE; and 4) deep multiple-class OSC methods: DTC. Abbreviations in parentheses. Specifically,
\begin{itemize}
	\item Iforest: an ensemble tree method to detect outliers;
	\item One-Class SVM (One-SVM): a baseline for out-of-class detection and classification;
	\item LACU-SVM (LACU): a SVM-based method that incorporates the unlabeled data from open set for unknown class detection;
	\item SENC-MAS (SENC): a matrix sketching method that approximates original information with a dynamic low-dimensional structure;
	\item ODIN-CNN (ODIN): a CNN-based method that distinguishes in-distribution and out-of-distribution over softmax score;
	\item CFO: a generative method that adopts an encoder-decoder GAN to generate synthetic unknown instances;
	\item CPE: a CNN-based ensemble method, which adaptively updates the prototype for detection;
%	\item MCL: a meta classification method that leverages pairwise similarity between instances;
	\item DTC: a extended deep transfer clustering method for novel class detection.
\end{itemize}
There are several instructions for baselines: 1) Iforest, ODIN and CFO can only perform binary classifications, i.e., whether the instance is an unknown class. Thus we further conduct unsupervised clustering on both know and unknown class data for subdividing; 2) all baselines are one-class methods except DTC, i.e., they also perform OSC in two steps: first detect the super-class of unknown classes, then perform unsupervised clustering; 3) all of baselines are OSC methods except LACU, SENC and CPE, but they can be applied in incremental OSC by combing memory data to update following~\cite{WangKCTK19}, except Iforest which replies on the quality of clustering in current time window.

%The main reason is that, for incremental OSC, unknown class detection can be quickly performed online, and then subdivided offline. Yet incremental clustering is more time-consuming and less effective

%, besides, we divide 10$\%$ of the training data as validation set for early stopping
\subsection{Implementation}
We develop $f$ based on convolutional network structure ResNet34~\cite{HeZRS16}, and $g$ based on ResNet18~\cite{HeZRS16}. Note that we use an identical set of hyperparameters ($\lambda= 1$, $\alpha = 0.3$, $\lambda_u= 0.2$, $\tau = 0.85$, $T (softmax-T) = 3$, $M = 2000$). In all of our models and experiments, we adopt standard SGD with Nesterov momentum~\cite{SutskeverMDH13}, where the momentum $\beta=0.9$. We train the initial model $f$ as following: the number of epochs is 20, the batch size is 64, the learning rate is 0.01, and weight decay is 0.001. While training $g$ as following: the number of epochs is 30, the batch size is 64, the learning rate is 0.005, and weight decay is 0.0005. We implement all baselines and perform all experiments based on code released by corresponding authors. For CNN based methods, we use the same network architecture and parameters during training, such as optimizer, learning rate schedule, and data pre-processing. Our method is implemented on a Nvidia TITAN X GPU with Pytorch 0.4.06~\footnote{https://pytorch.org/}.

%The output layer size is also 512.  in Python 3.6.2

\subsection{Open Set Classification}
We additionally validate the effectiveness of S2OSC on various settings: 1) ablation study about S2OSC (section \ref{sec:s5}); 2) performance with different number of unknown classes (section \ref{sec:s1}); 3) case study of filtered out-of-class instances (section \ref{sec:s2}). Considering page limitation and representativeness of used datasets, we utilize three typical datasets, i.e., MNIST, CIFAR-10 and SVNH, for additional experiments.
%4) performance on dataset with large-scare classes (section \ref{sec:s3}).

\begin{table*}[t]{\small
		\centering
		\caption{Ablation study about variants of S2OSC.}
		\label{tab:tab1}
		\begin{tabular*}{1\textwidth}{@{\extracolsep{\fill}}@{}l|c|c|c|c|c|c}
			\toprule
			\multirow{2}{*}{Methods} & \multicolumn{3}{c|}{Accuracy } & \multicolumn{3}{c}{F1 }\\
			\cmidrule(l){2-7}
			& MNIST  & CIFAR-10 & SVHN & MNIST  & CIFAR-10 & SVHN\\
			\midrule
			S2OSC-S &.834 &.812 &.748  &.792 &.771 &.666  \\
			S2OSC-LM &.793 &.675 &.589  &.754 &.608 &.505\\
			S2OSC-FN &.891 &.817 &.880  &.846 &.777 &.834\\
			S2OSC-U &.886 &.786 &.812  &.853 &.813 &.833\\
			S2OSC  &\bf.985 &\bf.847 &\bf.898  &\bf.985  &\bf.854  &\bf.901 \\
		\end{tabular*}
		\begin{tabular*}{1\textwidth}{@{\extracolsep{\fill}}@{}l|c|c|c|c|c|c}
			\toprule
			\multirow{2}{*}{Methods} & \multicolumn{3}{c|}{Precision} & \multicolumn{3}{c}{Recall}\\
			\cmidrule(l){2-7}
			& MNIST  & CIFAR-10 & SVHN & MNIST  & CIFAR-10 & SVHN\\
			\midrule
			S2OSC-S &.758 &.736 &.602  &.834 &.812 &.748\\
			S2OSC-LM &.727 &.561 &.465  &.793 &.675 &.589\\
			S2OSC-FN &.807 &.746 &.795  &.891 &.817 &.880 \\
			S2OSC-U &.765 &.758 &.876  &.786 &.886 &.812 \\
			S2OSC & \bf.847  & \bf.972 & \bf.888  &\bf .799 & \bf.986 &\bf .985\\
			\bottomrule
	\end{tabular*}}
\end{table*}

\begin{figure}[htb]
	\begin{center}
		\begin{minipage}[h]{48mm}
			\centering
			\includegraphics[width=48mm]{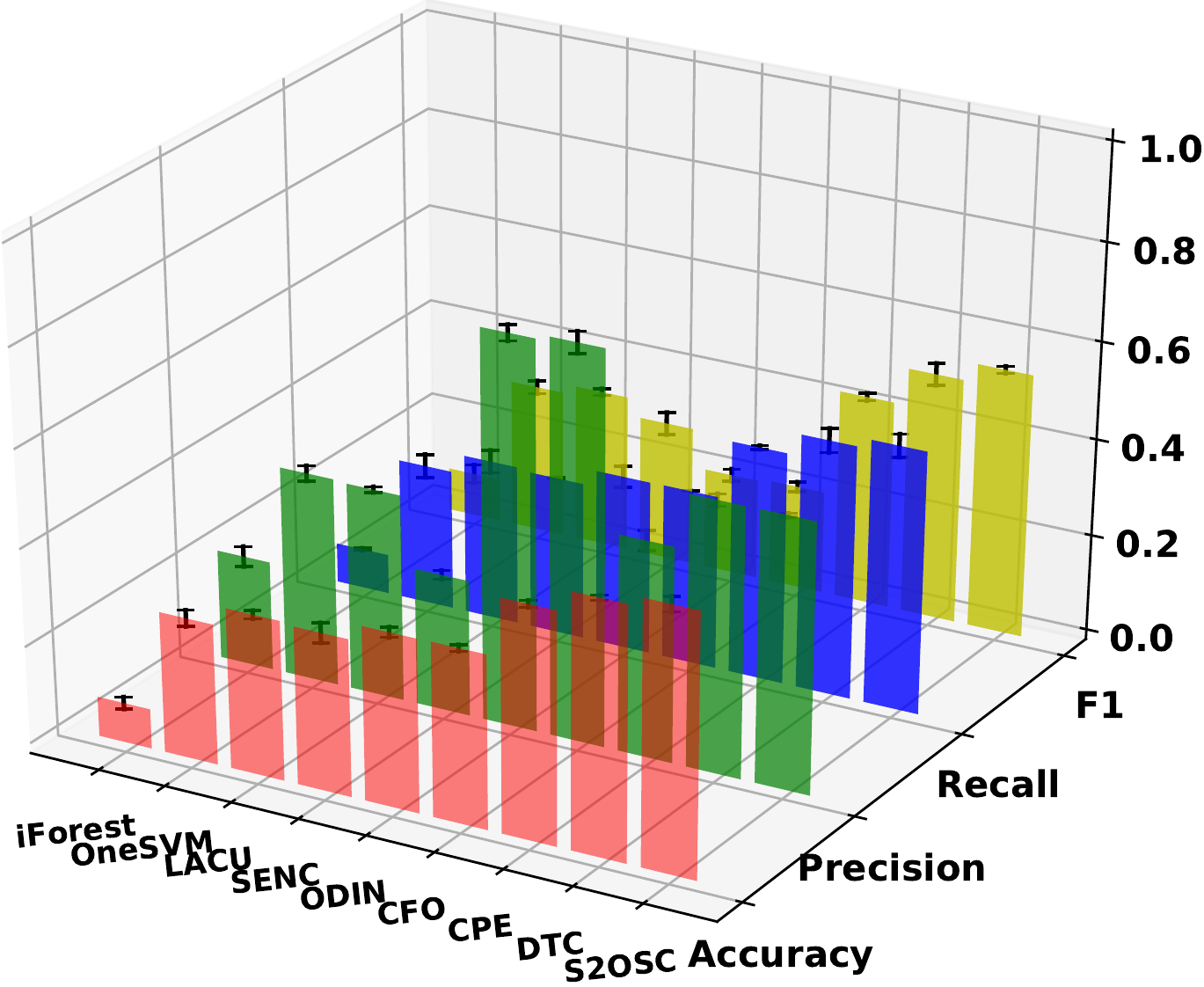}\\
			\mbox{  ({\it a}) {MNIST-3}}
		\end{minipage}
		\begin{minipage}[h]{48mm}
			\centering
			\includegraphics[width=48mm]{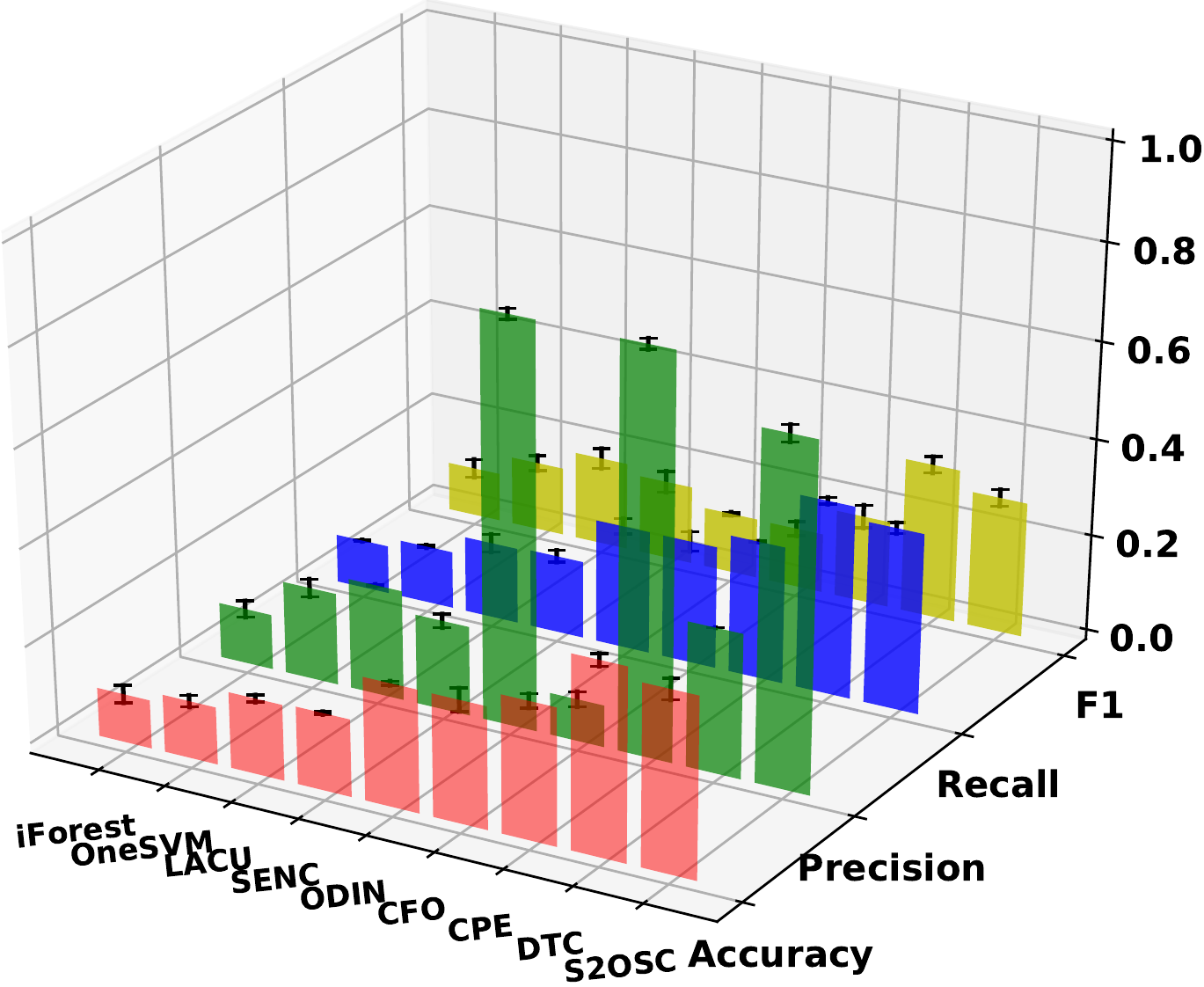}\\
			\mbox{  ({\it b}) {MNIST-5}}
		\end{minipage}\\
		\begin{minipage}[h]{48mm}
			\centering
			\includegraphics[width=48mm]{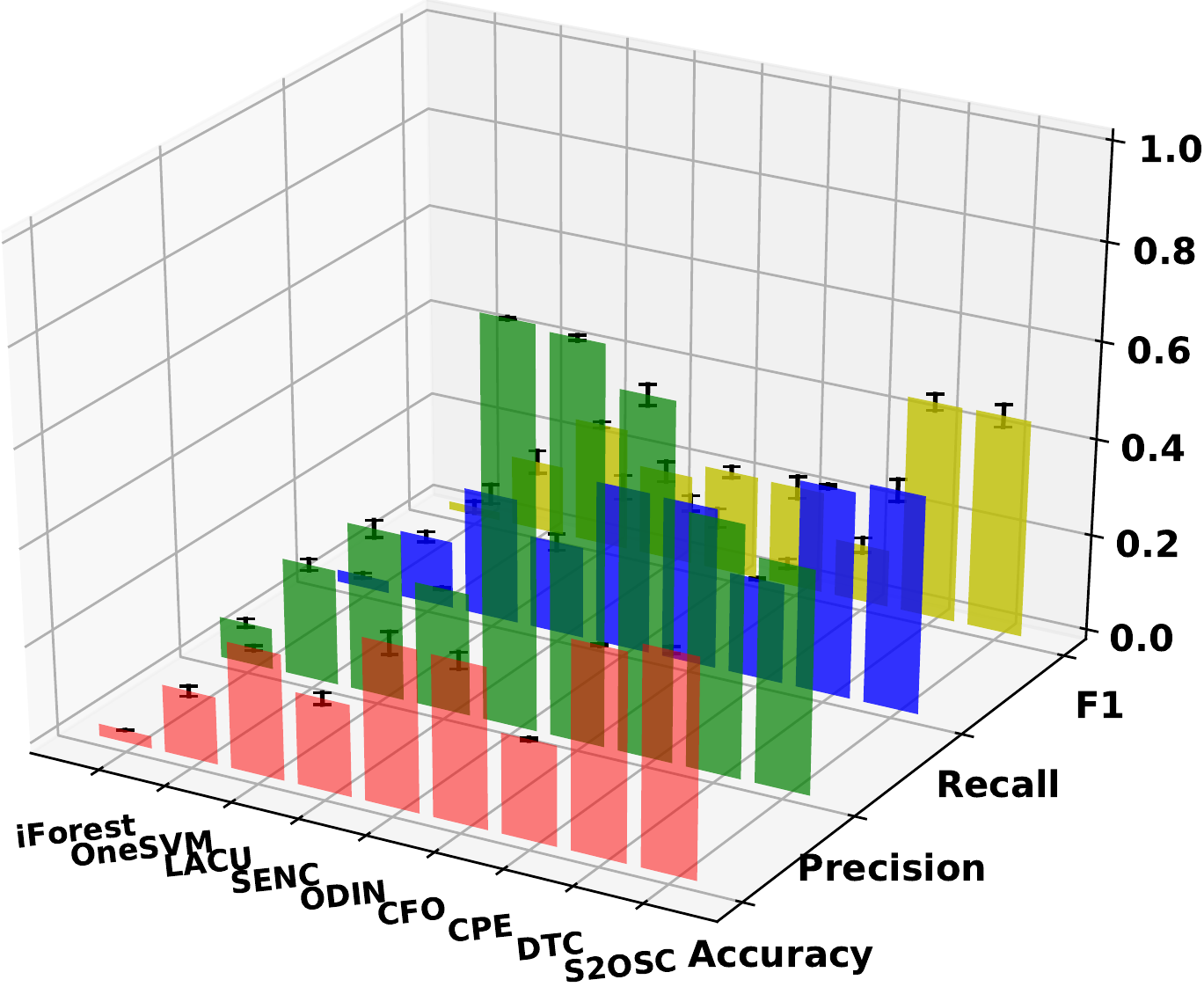}\\
			\mbox{  ({\it c}) {CIFAR-3}}
		\end{minipage}
		\begin{minipage}[h]{48mm}
			\centering
			\includegraphics[width=48mm]{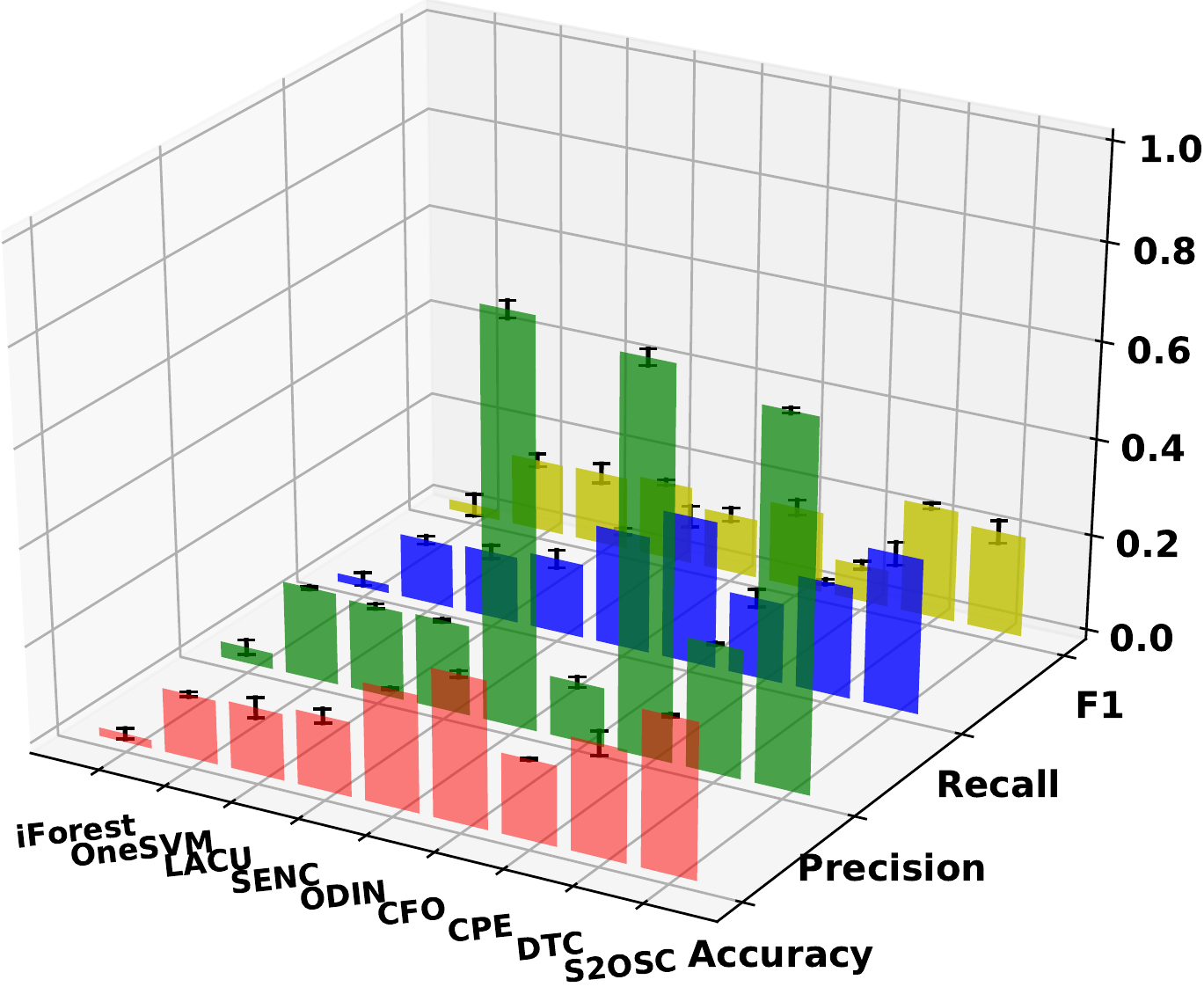}\\
			\mbox{  ({\it d}) {CIFAR-5}}
		\end{minipage}\\
		\begin{minipage}[h]{48mm}
			\centering
			\includegraphics[width=48mm]{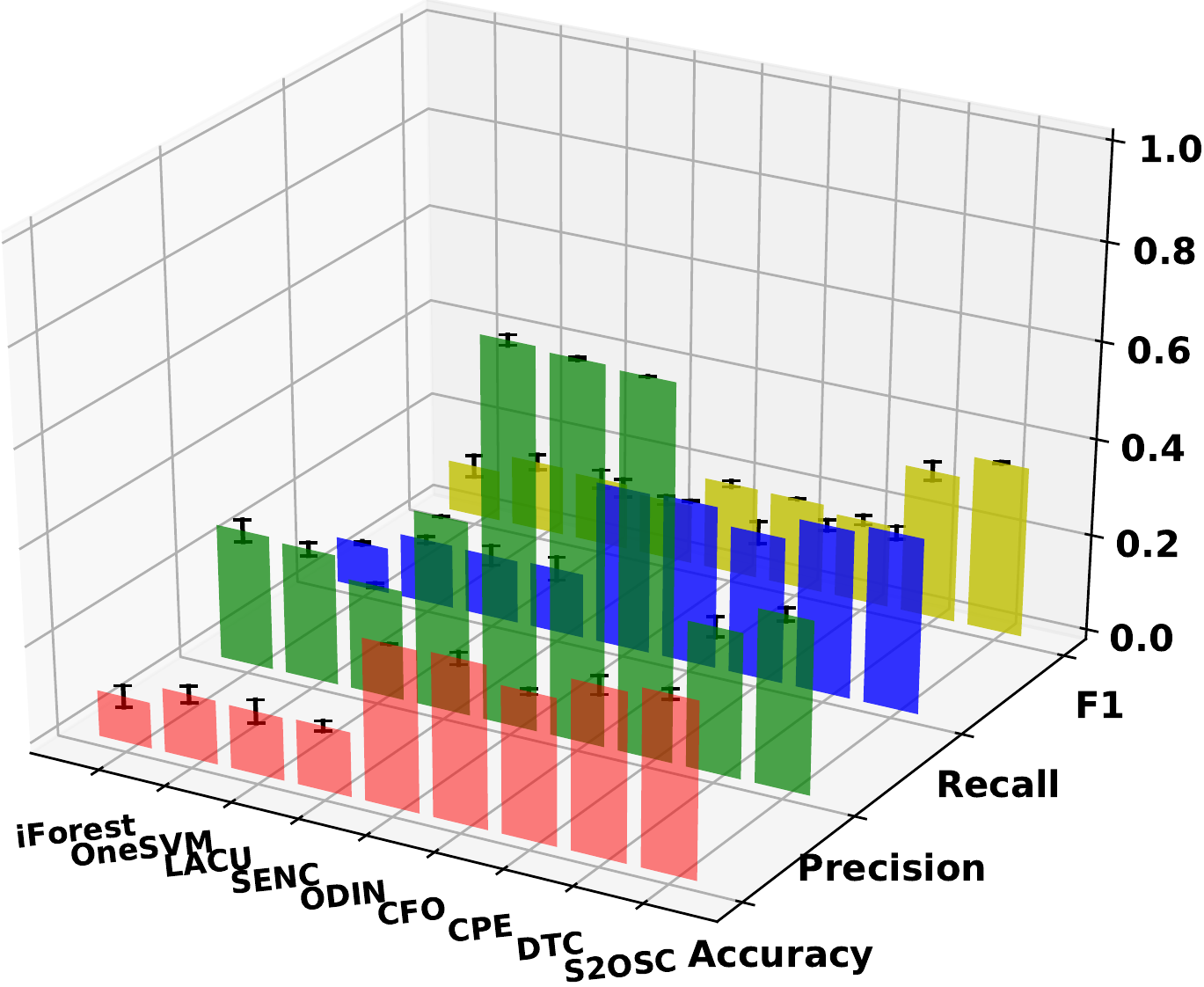}\\
			\mbox{  ({\it e}) {SVNH-3}}
		\end{minipage}
		\begin{minipage}[h]{48mm}
			\centering
			\includegraphics[width=48mm]{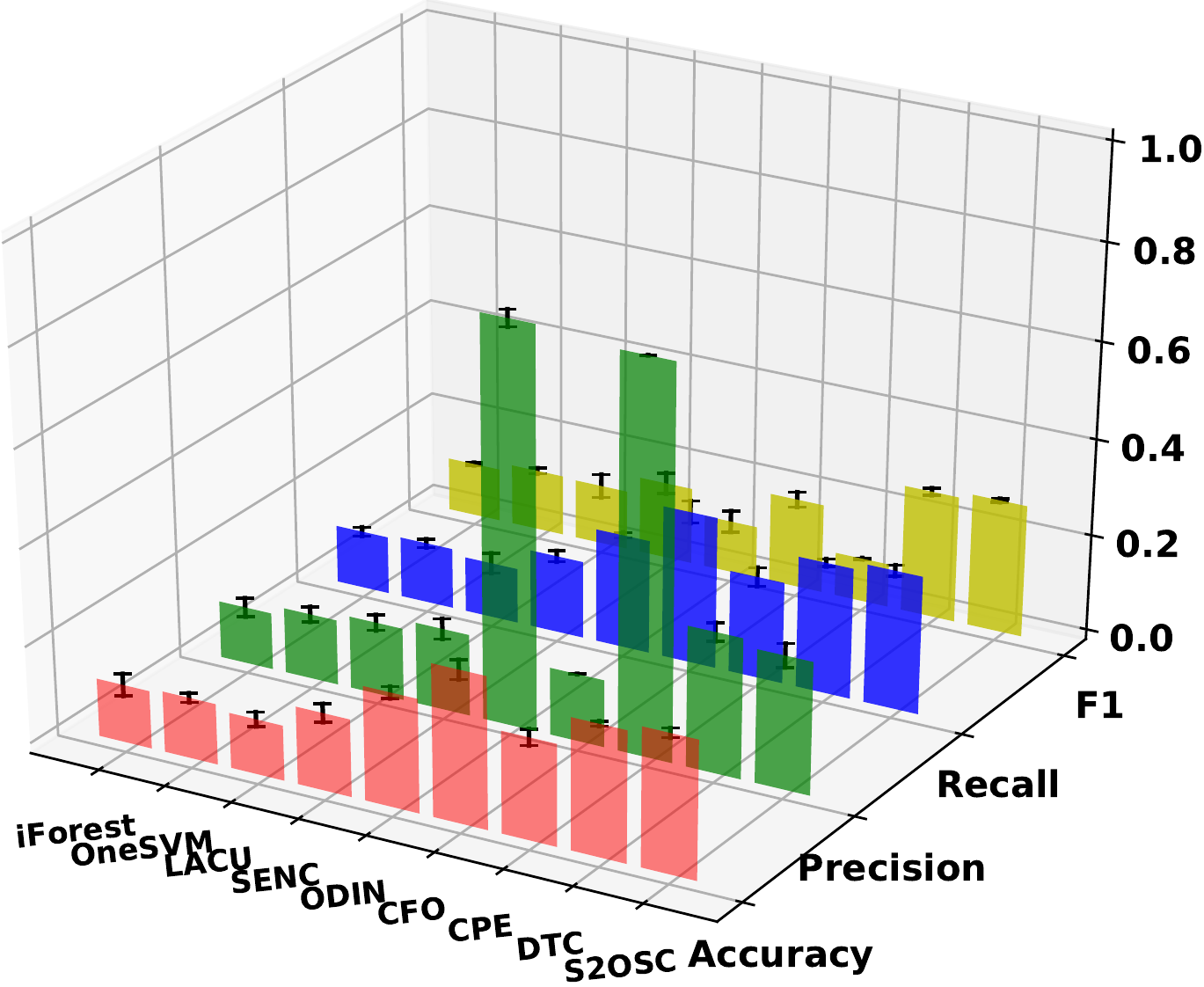}\\
			\mbox{  ({\it f}) {SVNH-5}}
		\end{minipage}\\
	\end{center}
	\caption{Classification performance of S2OSC and baseline methods on MNIST, CIFAR-10 and SVNH. Dataset-$p$ denotes with $p$ unknown classes. The X axis represents different OSC methods, the Y axis denotes the metrics, and the Z axis gives the performance.}\label{fig:f1}
\end{figure}

\subsubsection{Ablation Study}\label{sec:s5}
S2OSC has several variant designs. Therefore, in this subsection, we aim to analyze following questions: 1) Why choose the cross-entropy loss function for training $g$, without considering the large margin based loss function in traditional OSC methods~\cite{DaYZ14,WangKCTK19,geng2020}, such as triplet loss~\cite{Kulis13}, largin margin softmax~\cite{LiuWYY16}? 2) Why not directly perform semi-supervised learning based on the pre-trained model $f$ to obtain $g$, i.e., fine-tune the $f$ to acquire $g$? 3) Why not utilize only labeled data to calculate supervised loss for training $g$?

To answer the first question, traditional OSC methods adopted large margin loss to restrict intra-class and inter-class distance property, then detected novel class by identifying outliers. All of these approaches have a strong assumption that the embeddings or predictions learned by pre-trained model for out-of-class instances are apart from in-class ones. However, this assumption will fail on the data with complex content. Therefore, there is no need to use large margin loss here, and a unified classifier $f$ is more convenient for knowledge distillation in training $g$. More importantly, in our semi-supervised paradigm, for training classifier $g$, we first unify all unknown classes in $\D_{te}$ as a super-class. If we utilize large margin loss function here, it will also reduce the intra-class distance of super-class, which further affect $g$'s training considering semantic confusion and subsequent clustering operation. 

As to the second question, the model $f$ is pre-trained with in-class data, but performing semi-supervised re-training based on the pre-trained model $f$ has two limitations: 1) the number $K$ (examples in each class) of $\X$ is much smaller than that of $f$. Therefore, it is more inclined to classify the known classes and ignore the unknown classes if we fine-tune based on $f$; 2) the $f$ can not be regarded as teacher network for knowledge distillation any more.

For last problem, we train $g$ in the semi-supervised manner as~\cite{Kihyuk2020}, which achieves the state-of-the-art classification performance comparing with other approaches. The number $K$ of examples in each class is limited, and thus supervised training may leads to overfitting. 

%In \cite{Kihyuk2020}, weakly augmentation only conduct a standard flip-and-shift augmentation on the raw data, which can effectively retain the information of raw data. Thus, it only calculates the supervised loss on weakly augmented data, rather than combines raw data for training. 

To further verify our analysis, we conducted extra ablation experiments with the same setting as S2OSC on one unknown class: 1) \textbf{S2OSC-S} represents that we calculate loss $L_u$ including strongly-augmented instances instead of weakly-augmented version; 2) \textbf{S2OSC-LM} denotes that S2OSC utilizes the common triplet loss to train $g$; 3) \textbf{S2OSC-FN} indicates that S2OSC directly fine-tunes based on $f$ to obtain $g$; 4) \textbf{S2OSC-U} represents that S2OSC removes the unsupervised term. Table \ref{tab:tab1} records the results, best results are in bold. We observe that S2OSC outperforms all variant methods on three datasets with different criteria. This validate that: 1) the performances of strong  augmentation is worse than weak augmentation, this indicates that the strong augmentation may introduce more noises which are difficult to train; 2) large-margin loss is not suitable for training $g$, even only with one unknown class; 3) fine-tuning base on $f$ is not a good choice, which validates the analysis above; 4) the contribution of unlabeled data is significant, this indicates that unlabeled data can enlarge the training data and contribute for the learning process.

%this may be due to the embedding confused examples still contained in pseudo-labeled unknown class data, which causes the label noise in utilizing unlabeled data. 
%Note the pseudo-labeled out-of-class data also include embedding confused examples, which may affects the training of $g$.
%, and is competitive with S2OSC-S and S2OSC-U
%. This is because that there exist label noise in $\X$, and hard margin may affect the training

%		\begin{tabular*}{1\textwidth}{@{\extracolsep{\fill}}@{}l|c|c|c|c|c|c|c|c|c|c}
%			\toprule
%			\multirow{2}{*}{Methods} & \multicolumn{5}{c|}{F$_{in}$} & \multicolumn{5}{c}{F$_{out}$}\\
%			\cmidrule(l){2-11}
%			& F-MN & CIFAR & SVHN & MN & CINIC & F-MN & CIFAR & SVHN & MN & CINIC\\
%			\midrule
%			Iforest  &.989 &.861 &.857  &.818 &.824 &.943 &.251  &.230 &.065  &.205\\
%			One-SVM  &.704 &.738 &.638  &.737 &.736 &.467 &.279  &.300 &.461  &.285\\
%			LACU     &.787 &.848 &.879  &.879 &.596 &.398 &.190  &.251 &.223  &.348\\
%			SENC     &.906 &.899  &.908 &.899 &.907 &.697  &.110 &.007 &.005  &.005\\
%			\midrule
%			ODIN     &.954 &\bf.909 &.953  &.954 &.909 &.391 &.031  &.219 &.390  &-\\
%			CFO      &.905 &.905 &.849  &.843 &.849 &.275 &.275  &.285 &.278  &.285\\
%			CPE &.871 &.825 &.841  &.988 &.814 &.382 &.383  &.563 &.923  &.239 \\
%			\midrule
%%			MCL  &.920 &\bf.920 &.880  &.970 &881 &- &-  &- &-  &.052\\
%			DTC &.908 &.805 &.935  &.858 &\bf.984 &.212 &.252  &.142 &.249  &.353\\
%			\midrule
%			S2OSC  &\bf.999 & .882 &\bf.938  &\bf.988 &.848 &\bf.993 &\bf.853  &\bf.932 &\bf.987  &\bf.827\\
%		\end{tabular*}

\begin{table*}[t]\scriptsize
		\centering
		\caption{Comparison of novel class detection (F$_{out}$) with different number of unknown classes (SVM/Our denotes One-SVM/S2OSC). }
		\label{tab:tab2}
		\resizebox{\textwidth}{!}{
		\begin{tabular*}{1\textwidth}{@{\extracolsep{\fill}}@{}l|@{}c|@{}c|@{}c|@{}c|@{}c|@{}c|@{}c|@{}c|@{}c}
			\toprule
			 & \multicolumn{3}{c|}{One Class} & \multicolumn{3}{c|}{Three Classes} &\multicolumn{3}{c}{Five Classes}\\
			\cmidrule(l){2-10}
			& MNIST  & CIFAR & SVHN & MNIST  & CIFAR & SVHN & MNIST  & CIFAR & SVHN\\
			\midrule
			Iforest  &.065$\pm$.010 &.251$\pm$.032  &.230$\pm$.021  &.292$\pm$.047 &.139$\pm$.041 &.224$\pm$.037 &.407$\pm$.060 &.224$\pm$.031 &.219$\pm$.014
			\\
			SVM  &.461$\pm$.035 &.279$\pm$.016  &.300$\pm$.028  &.494$\pm$.093 &.525$\pm$.057 &.519$\pm$.060 &.488$\pm$.057 &.479$\pm$.067 &.464$\pm$.044
			\\
			LACU     &.223$\pm$.026 &.190$\pm$.009  &.251$\pm$.018  &.381$\pm$.058 &.449$\pm$.026 &.465$\pm$.102 &.514$\pm$.067 &.604$\pm$.036	&.594$\pm$.019
			\\
			SENC     &.005$\pm$.000 &.110$\pm$.003 &.007$\pm$0  &.530$\pm$.002  &.554$\pm$.040 &.472$\pm$.075 &.690$\pm$.004 &.689$\pm$.003 &.706$\pm$.018
			\\
			\midrule
			ODIN     &.390$\pm$.085 &.031$\pm$0  &.219$\pm$.010  &.006$\pm$0 &.006$\pm$0 &.006$\pm$0 &.130$\pm$0 &.130$\pm$0	&.130$\pm$0
			\\
			CFO      &.278$\pm$.013 &.275$\pm$.017  &.285$\pm$.012  &.296$\pm$.110 &.442$\pm$.036 &.452$\pm$.096 &.543$\pm$.002 &.556$\pm$.010 &.596$\pm$.072
			\\
			CPE      &.923$\pm$.015 &.383$\pm$.022 &.563$\pm$.053  &.424$\pm$.055 &.385$\pm$0.026 &.338$\pm$.094 &.619$\pm$.040 &.557$\pm$.050 &.577$\pm$.119
			\\
			\midrule
			DTC      &.249$\pm$.041 &.252$\pm$.028  &.142$\pm$.016  &.074$\pm$.009  &.241$\pm$0.176 &.309$\pm$.084 &.553$\pm$.131 &.593$\pm$.002 &.756$\pm$.055
			\\
			\midrule
			Our    &\bf.987$\pm$.019 &\bf.853$\pm$.083  &\bf.932$\pm$.032  &\bf.629$\pm$013 &\bf.632$\pm$0.024 &\bf.666$\pm$.117 &\bf.974$\pm$.001 &\bf.858$\pm$.028 &\bf.830$\pm$.029\\
			\bottomrule
		\end{tabular*}}
\end{table*}

\begin{figure}[htb]
	\begin{center}
		\begin{minipage}[h]{33mm}
			\centering
			\includegraphics[width=33mm]{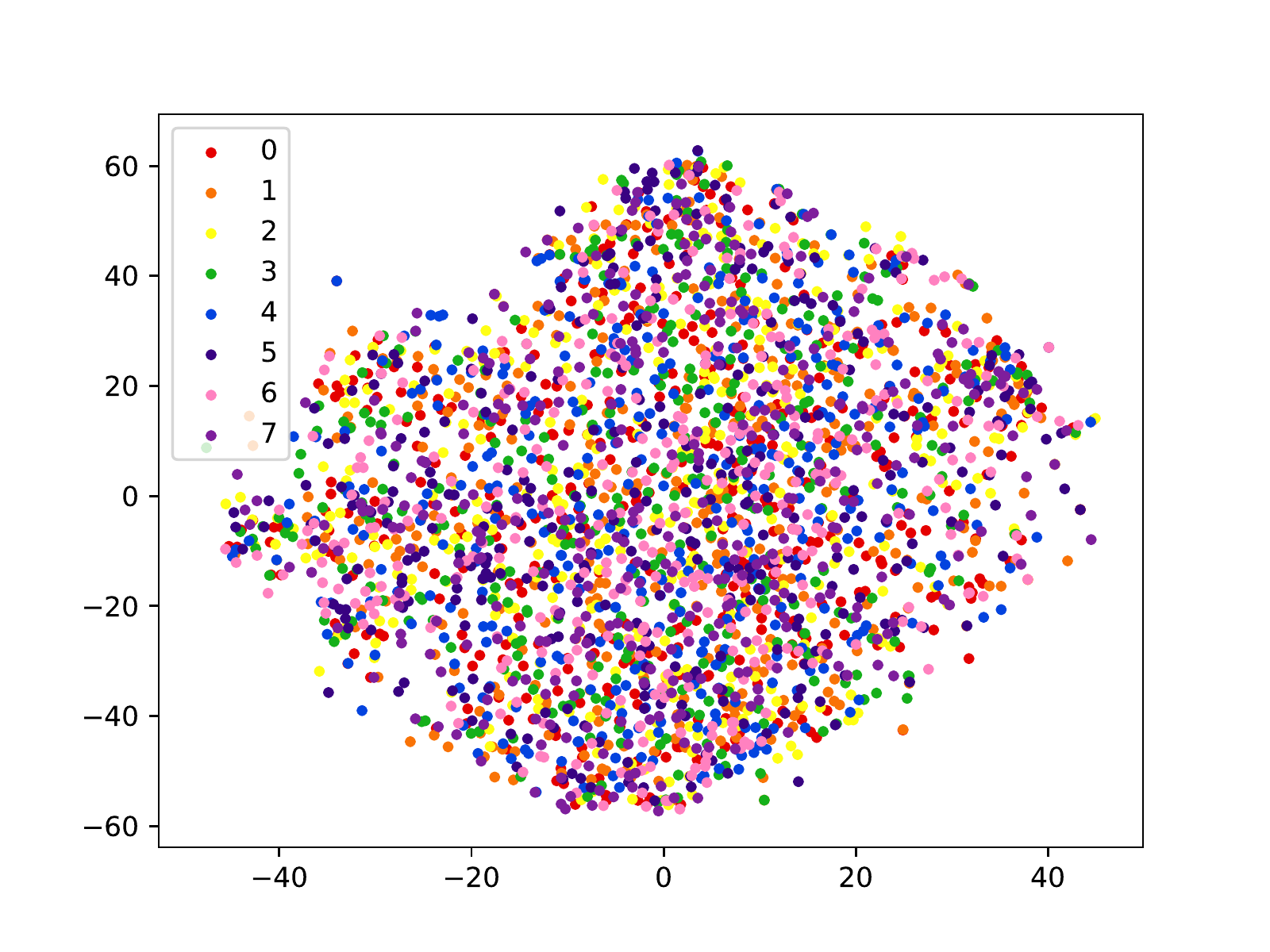}\\
			\mbox{  ({\it a-1}) {Original-3}}
		\end{minipage}
		\begin{minipage}[h]{33mm}
			\centering
			\includegraphics[width=33mm]{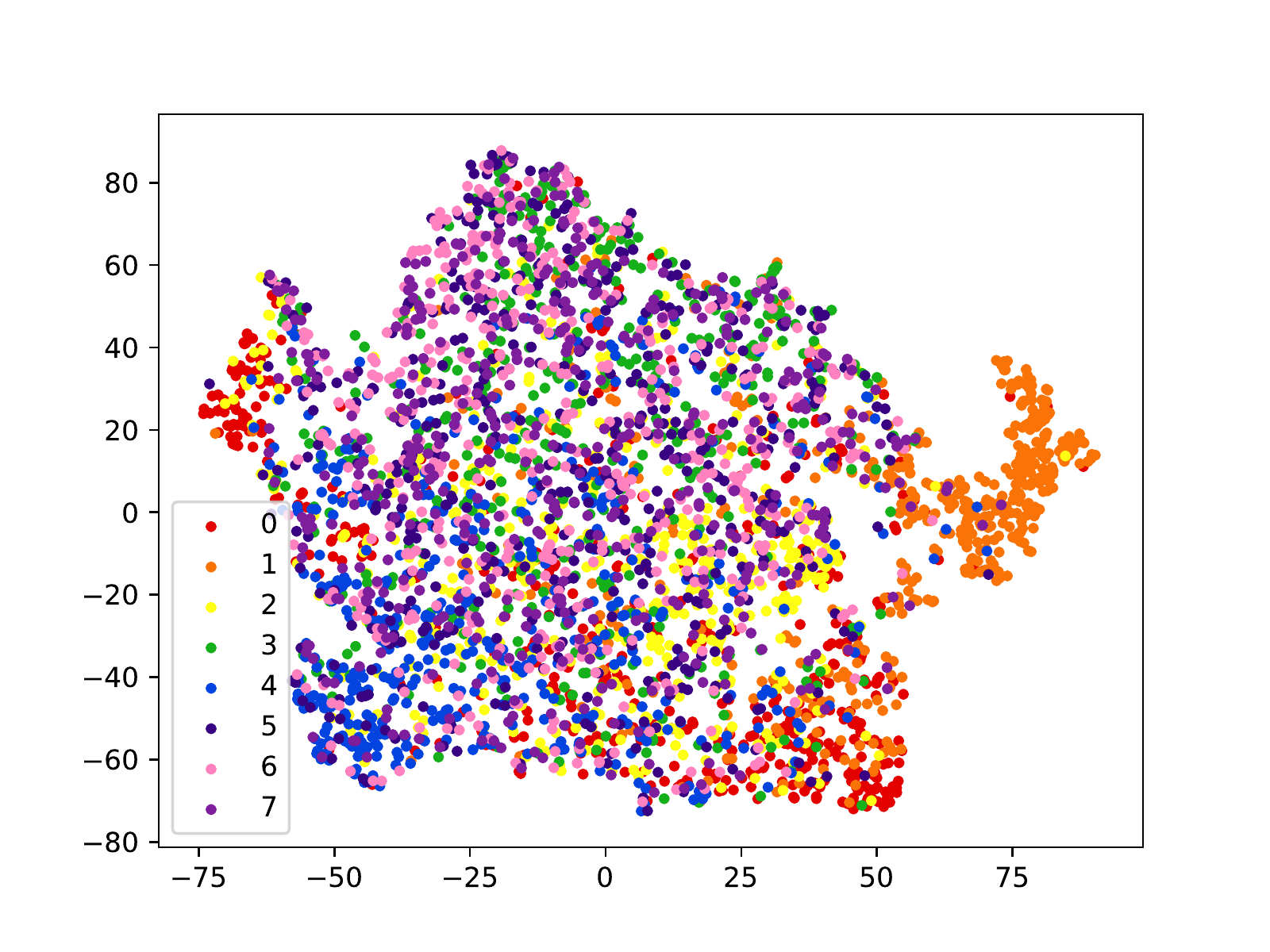}\\
			\mbox{  ({\it b-1}) {CPE-3}}
		\end{minipage}
		\begin{minipage}[h]{33mm}
			\centering
			\includegraphics[width=33mm]{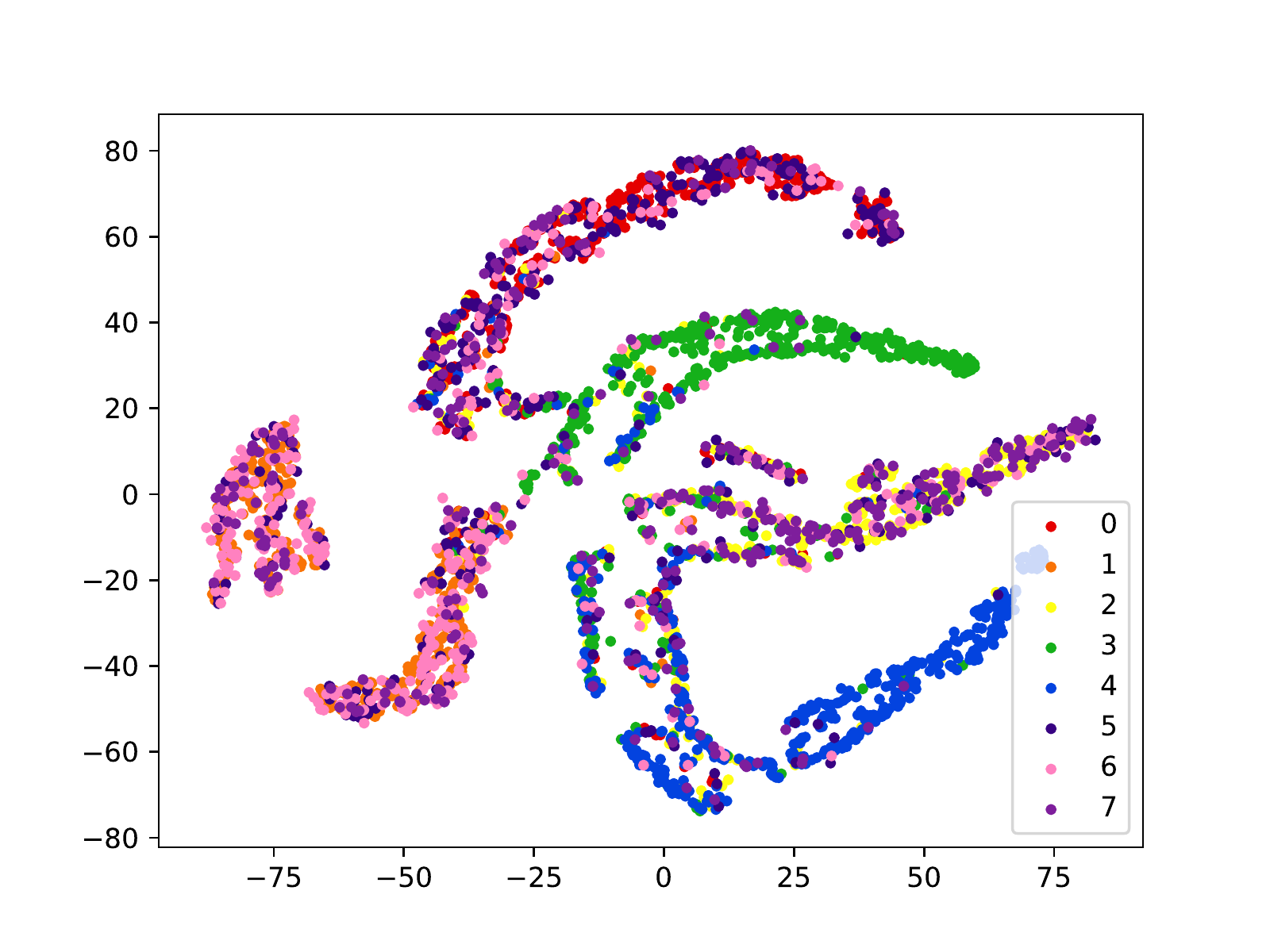}\\
			\mbox{ ({\it c-1}) {CFO-3}}
		\end{minipage}
		\begin{minipage}[h]{33mm}
			\centering
			\includegraphics[width=33mm]{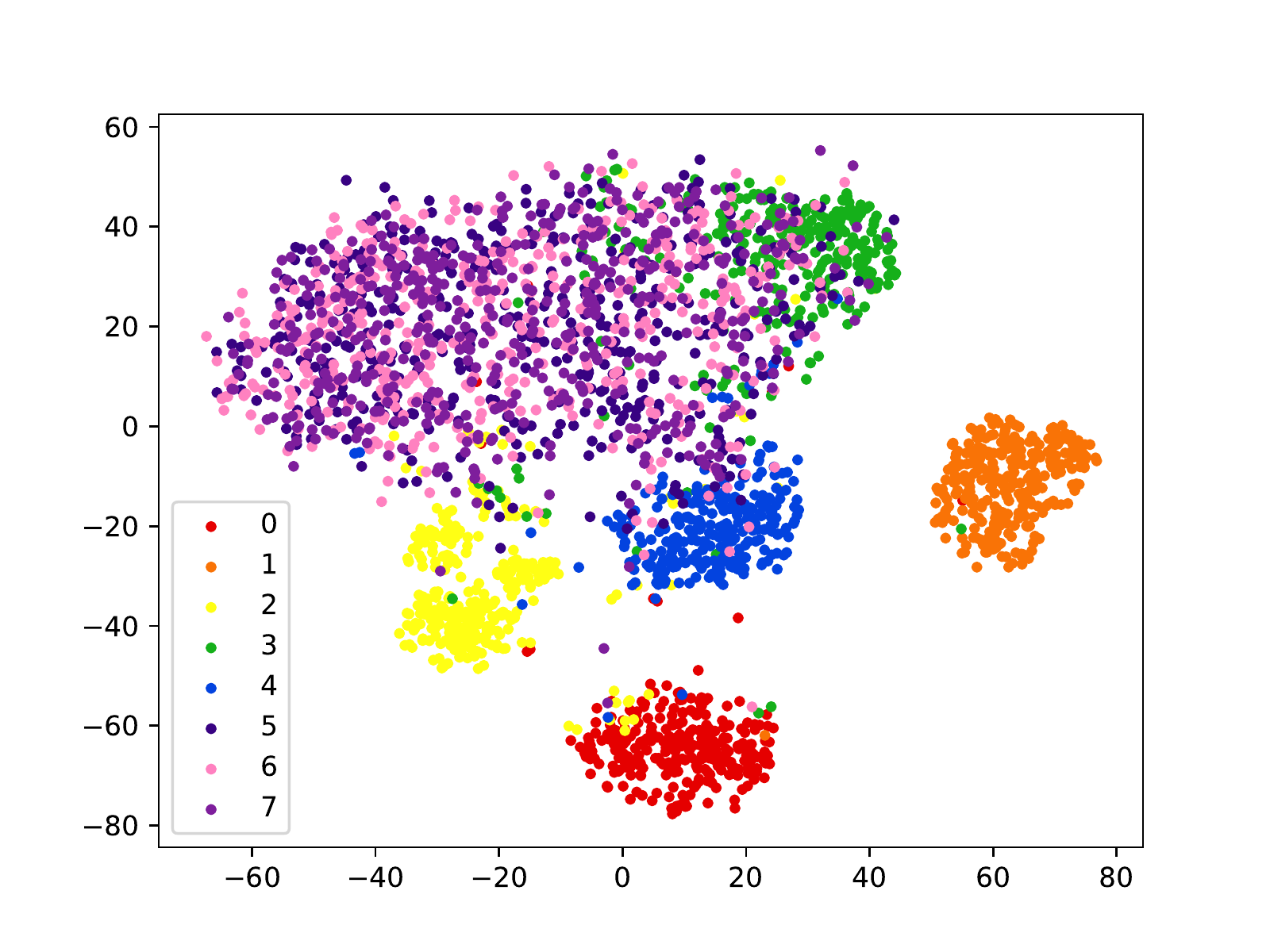}\\
			\mbox{ ({\it d-1}) {S2OSC-3}}
		\end{minipage}\\
		\begin{minipage}[h]{33mm}
			\centering
			\includegraphics[width=33mm]{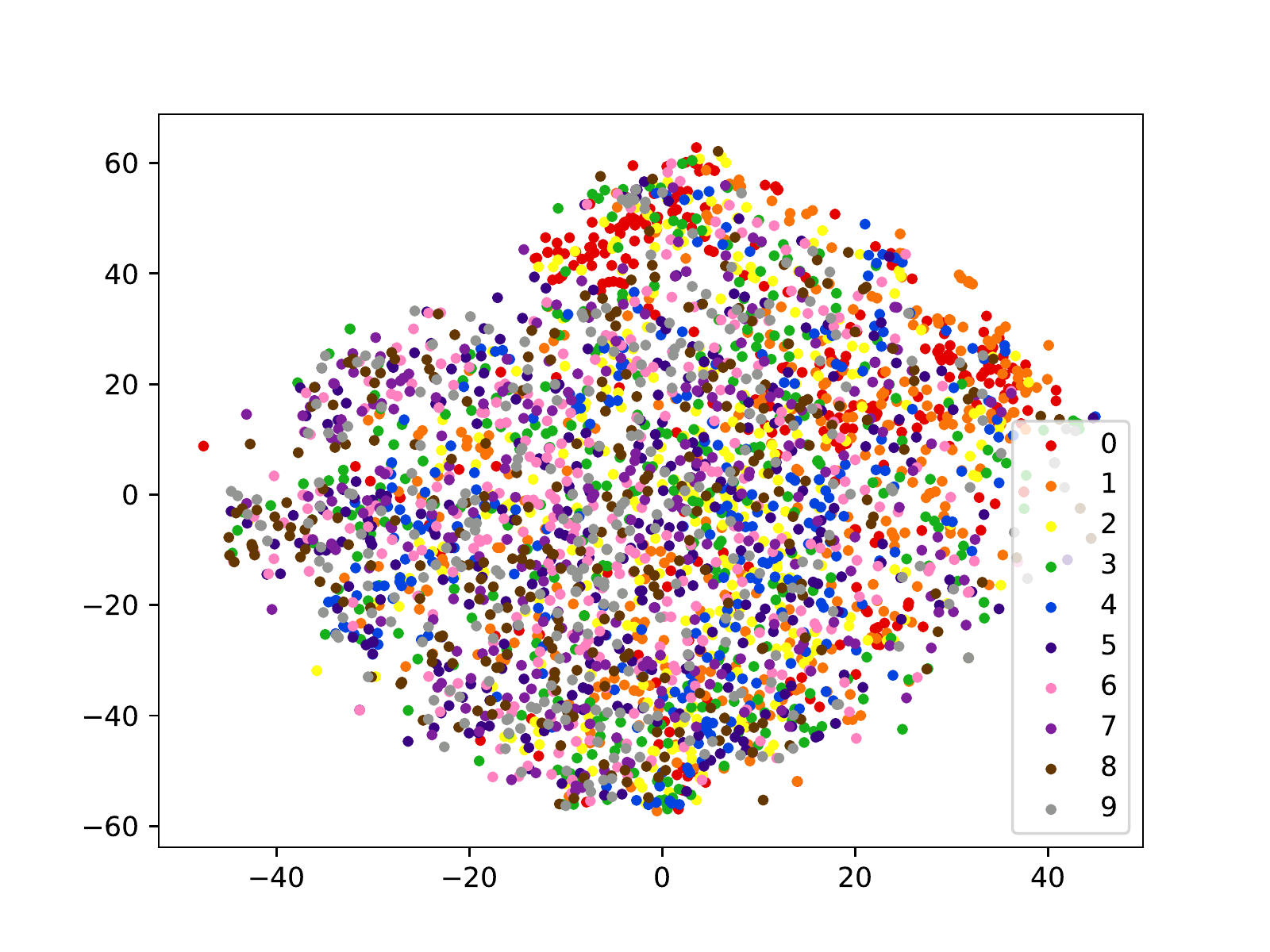}\\
			\mbox{  ({\it a-2}) {Original-5}}
		\end{minipage}
		\begin{minipage}[h]{33mm}
			\centering
			\includegraphics[width=33mm]{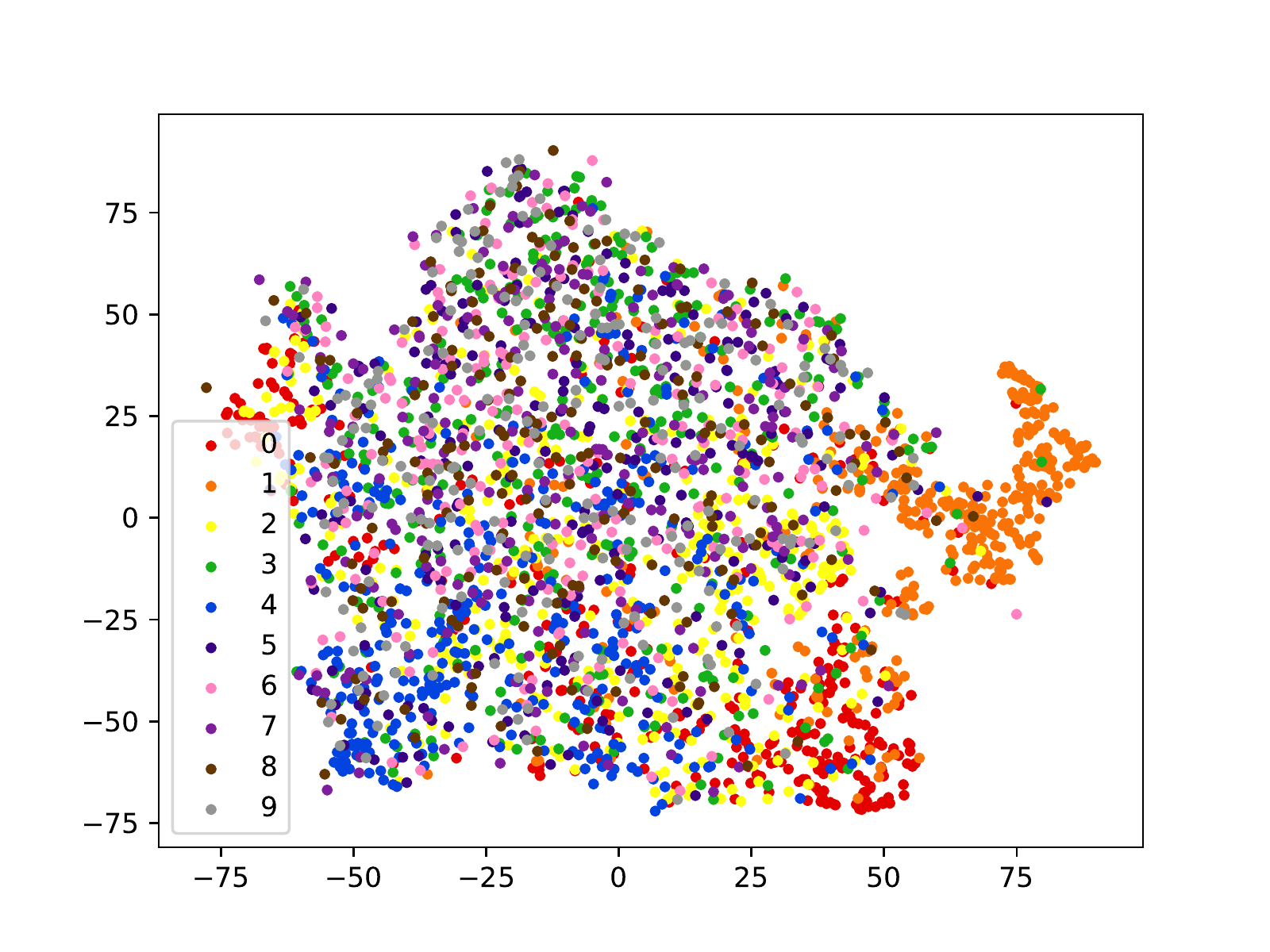}\\
			\mbox{  ({\it b-2}) {CPE-5}}
		\end{minipage}
		\begin{minipage}[h]{33mm}
			\centering
			\includegraphics[width=33mm]{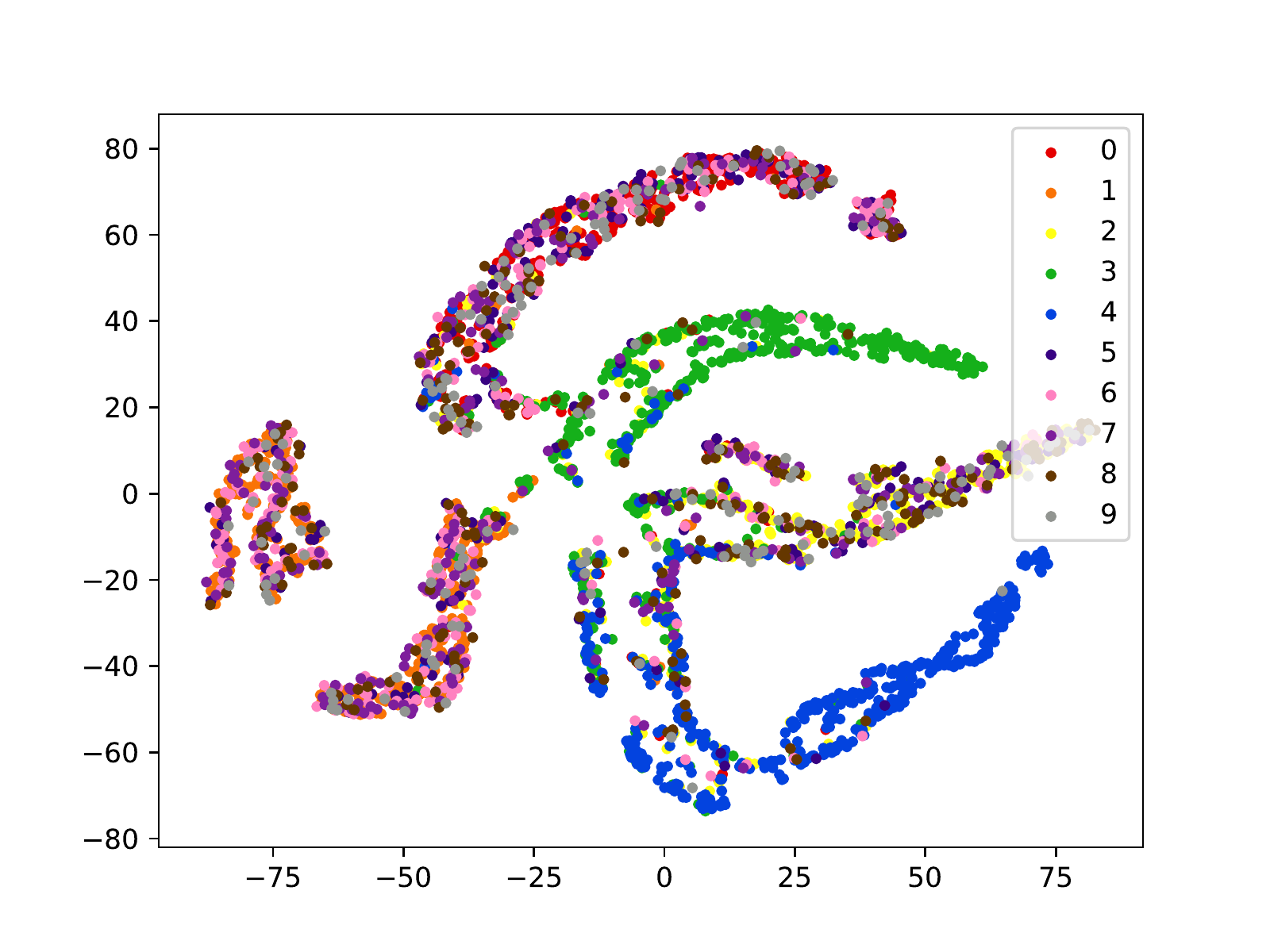}\\
			\mbox{ ({\it c-2}) {CFO-5}}
		\end{minipage}
		\begin{minipage}[h]{33mm}
			\centering
			\includegraphics[width=33mm]{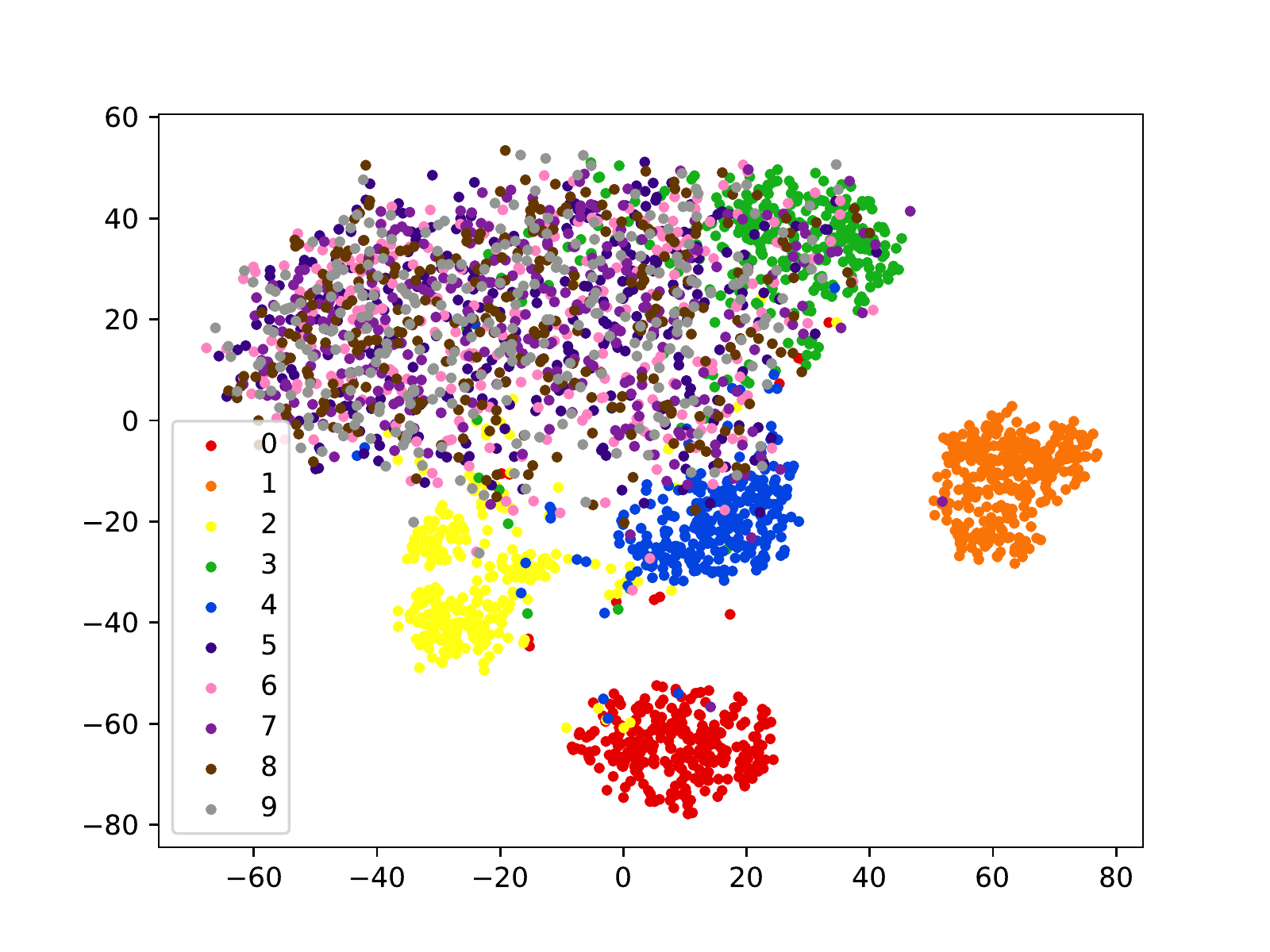}\\
			\mbox{ ({\it d-2}) {S2OSC-5}}
		\end{minipage}
	\end{center}
	\caption{T-SNE Visualization on CIFAR-10 dataset. Method-$p$ denotes with $p$ unknown classes. (a) original feature space; (b) learned representations by discriminative method CPE; (c) learned representations by generative detection method CFO; (d) learned representations by proposed S2OSC.}\label{fig:f2}
\end{figure}

\subsubsection{Influence of Unknown Class Number}\label{sec:s1}
Considering that all datasets have 10 classes, thus we randomly hold out 5 classes as initial training data, and additionally tune the number of unknown classes in $\{3, 5\}$ by randomly picking remaining classes (experiments about one unknown class have been reported, considering that our method is significantly superior to baselines, we did not report std). We also extract 33$\%$ of the known class data into test set, so that the test set mixes with known and unknown classes. The results report averaged performance over 5 random class partitions as~\cite{geng2020}.
%Note that we set $K=500$ for five unknown class

Figure \ref{fig:f1} records the experiment results (mean and std) of three typical datasets, which reveal that: 1) with the number of unknown classes increases, performance of all approaches decrease. This indicates that, once multiple unknown classes emerge in testing phase, the problem of embedding confusion will exacerbate, making OSC more complicated; 2) the precision of S2OSC is not high, whereas the recall of high precision model (e.g., CPE) is not very good, this indicates that most unknown classes are divided into known classes. The F$_{out}$ of novel class detection in Table  \ref{tab:tab2} also validates this phenomenon; 3) under three unknown classes scenario, S2OSC still outperforms all baselines on various criteria except recall. While S2OSC is competitive with other baselines under five unknown classes, i.e., S2OSC is lower than several baselines on MNIST dataset, and lower than CPE of precision on other two datasets.  

Figure \ref{fig:f2} shows feature embedding results using T-SNE~\cite{maaten2008visualizing}. The figures in upper row are T-SNE results with 3 unknown classes (i.e., 5, 6, 7), and figures in bottom row are with 5 unknown classes (i.e., 5, 6, 7, 8, 9, 10). Clearly, S2OSC can obviously distinguish between known and unknown classes, i.e., instances from unknown classes are well separated from  known clusters comparing with other baselines, which benefits unknown class detection in result. 
%CPE and CFO are difficult to distinguish between known and unknown classes in the case of multiple unknown classes, whereas

For further measuring the discrimination of known and unknown classes, we utilize another criterion in~\cite{WangKCTK19}, which treats OSC as a binary classification problem placing emphasis on novel class detection. In detail, we consider all known classes as negative and all unknown classes as positive. $F_{out} = \frac{2TP}{2TP+FP+FN}$ is F1 of unknown classes, $TP, FP, FN, TN$ denotes the true positives, false positive, false negatives and true negatives. Table \ref{tab:tab2} gives the results, best results are in bold. We can observe that novel class detection, i.e., F1 indicator, of S2OSC is significantly higher than other methods on various settings and ODIN is difficult to handle multiple unknown classes. An interesting phenomenon is that the F1 of three unknown classes is lower than that of one and five unknown classes. This is because one unknown class does not have the problem of intra-class confusion existing in multiple unknown classes, thus has better performance. On the other hand, with the number of unknown classes increasing, the embedding confused instances in filtered data decrease, thus it is more conducive to the training $g$ and improves performance.

\begin{figure*}[htb]\centering
	\centering
	\includegraphics[width = 140mm]{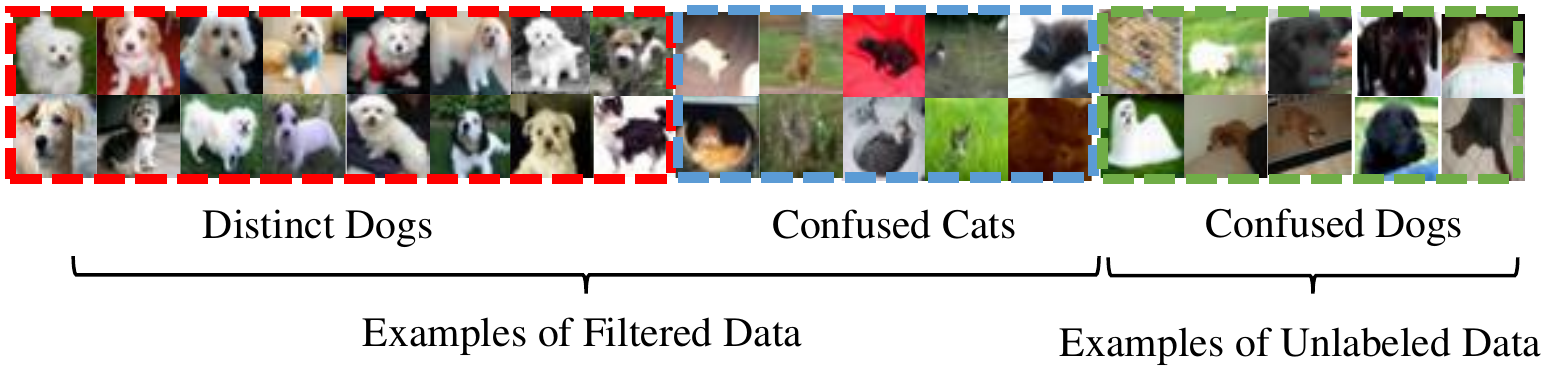}\\
	\caption{Examples of filtered instances with large weights $w$ and unselected out-of-class instances by proposed S2OSC.}\label{fig:f3}
\end{figure*}

\subsubsection{Case Study}\label{sec:s2}
We also give some examples of distinct and confused instances for display. Here we consider one unknown class case on CIFAR-10 dataset, in which the known class set includes: ``airplane, cat, deer, automobile, truck'', and the unknown class is ``dog''. Then we sort $\D_{te}$ according to the weights $w$, and select instances according to the ranking. As shown in Figure \ref{fig:f3}, we observe that: 1) most of the distinct instances are dogs, but still include few known class data, for example, instances from cat. It can be seen from the examples that many confused cats are outliers, which are difficult to distinguish; 2) most distinct dogs have more diagnostic characteristics, for example, the images with full-body shot; 3) unselected dogs are ambiguous, for example, dogs with only head or unclear dog images. Thus, combining labeled data into unlabeled loss is helpful for training $g$.

\subsubsection{Large-Scale OSC}\label{sec:s3}
To validate the effectiveness of our proposed method on large-scale OSC setting, i.e., dataset with large-scale classes. We conduct more experiments on CIFAR-30 following~\cite{geng2020}. In detail, we adopt the split class case following~\cite{geng2020}, which holds out 10 classes as out-of-class for testing, and leaves the remaining classes as the initial training set. The experimental setup is same as the case of multiple unknown classes. Table \ref{tab:tab6} records the results, best results are in bold. We can observe that: 1) the performances of all methods decrease rapidly facing large-scale OSC; 2) we have similar results as other setups that S2OSC consistently outperforms all baselines on various criteria except F1, which ranks runner-up. For example, S2OSC provides at least 10$\%$ improvements of accuracy than other baselines. This shows that S2OSC can well perform OSC on different class scales.

\begin{table}[htb]{
		\centering
		\caption{Comparison of large-scale open set classification with multiple unknown classes.}
		\label{tab:tab6}
		\begin{tabular*}{1\textwidth}{@{\extracolsep{\fill}}@{}l|c|c|c|c|c|c|c|c|c}
			\toprule
			Methods & Iforest & One-SVM & LACU & SENC & ODIN & CFO & CPE  & DTC & S2OSC\\
			\midrule
			Accuracy  &.006 &.009 &.008  &.007 &.163 &.147 &.176  &.181 &\bf.302   \\
			Precision &.709 &.703 &.195  &.179 &.340 &.436 &.245  &.248 &\bf.791   \\
			Recall    &.006 &.009 &.008  &.007 &.163 &.147 &.176  &.181 &\bf.302   \\
			F1        &.006 &.008 &.008  &.007 &.147 &.125 &.165  &\bf.168 &.165   \\
			F$_{out}$ &.649 &.614 &.691  &.706 &.917 &.898 &.917  &.917 &\bf.940   \\
			\bottomrule
	\end{tabular*}}
\end{table}

%first reserve test data from raw split as the hold-out set for each class. Then
\subsection{Incremental Open Set Classification}

\subsubsection{Streaming Dataset}
In this subsection, we mainly give the details of streaming data construction and measure criteria. In this paper, we perform incremental OSC, rather than online OSC, thus we need to accept the testing data before performing unknown class detection. There exists many related scenarios, such as sequential task learning in lifelong learning.

\begin{figure}[htb]
	\begin{center}
		\begin{minipage}[h]{60mm}
			\centering
			\includegraphics[width=60mm]{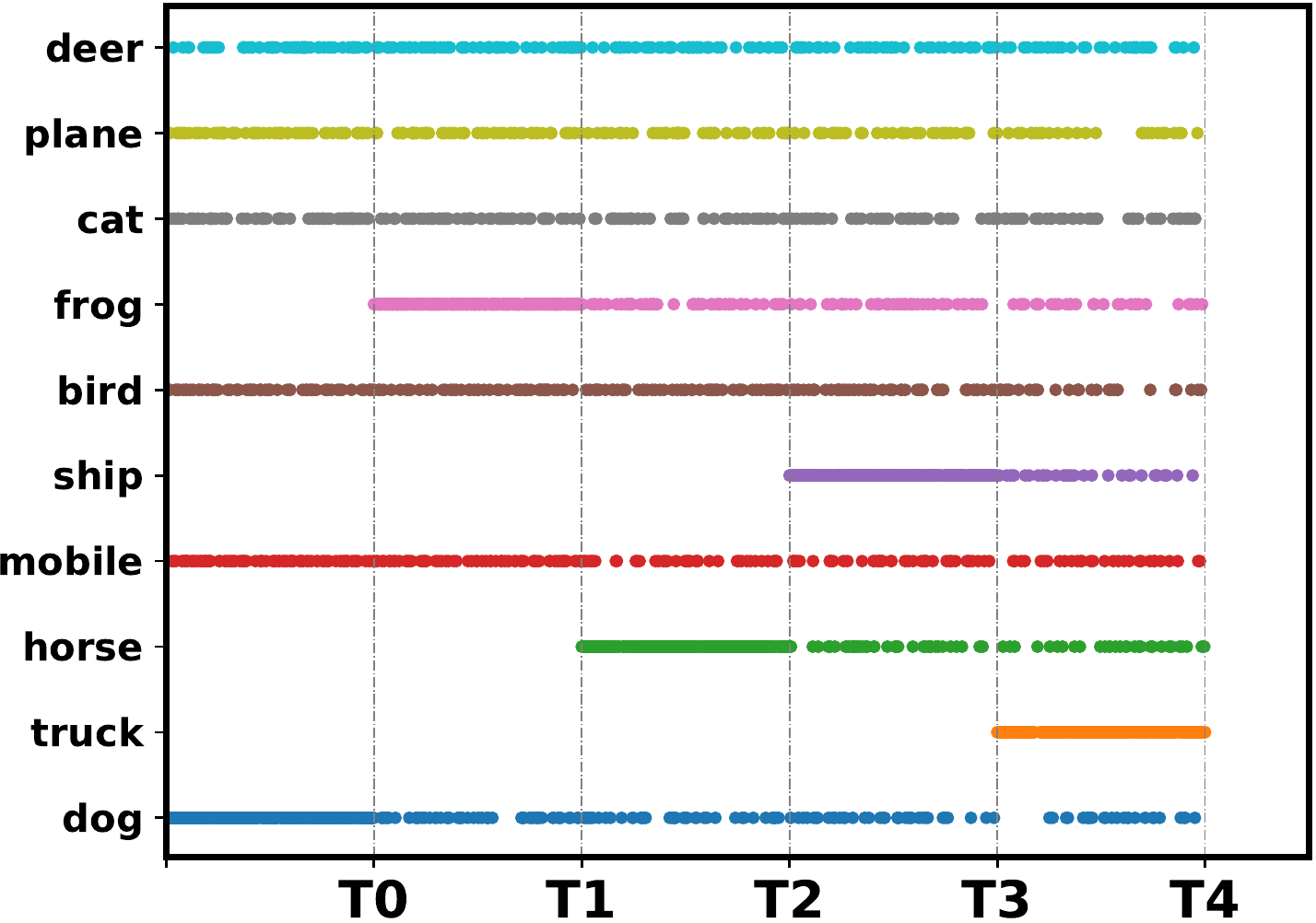}
			\mbox{  ({\it a}) {single unknown class}}
		\end{minipage}
		\begin{minipage}[h]{60mm}
			\centering
			\includegraphics[width=60mm]{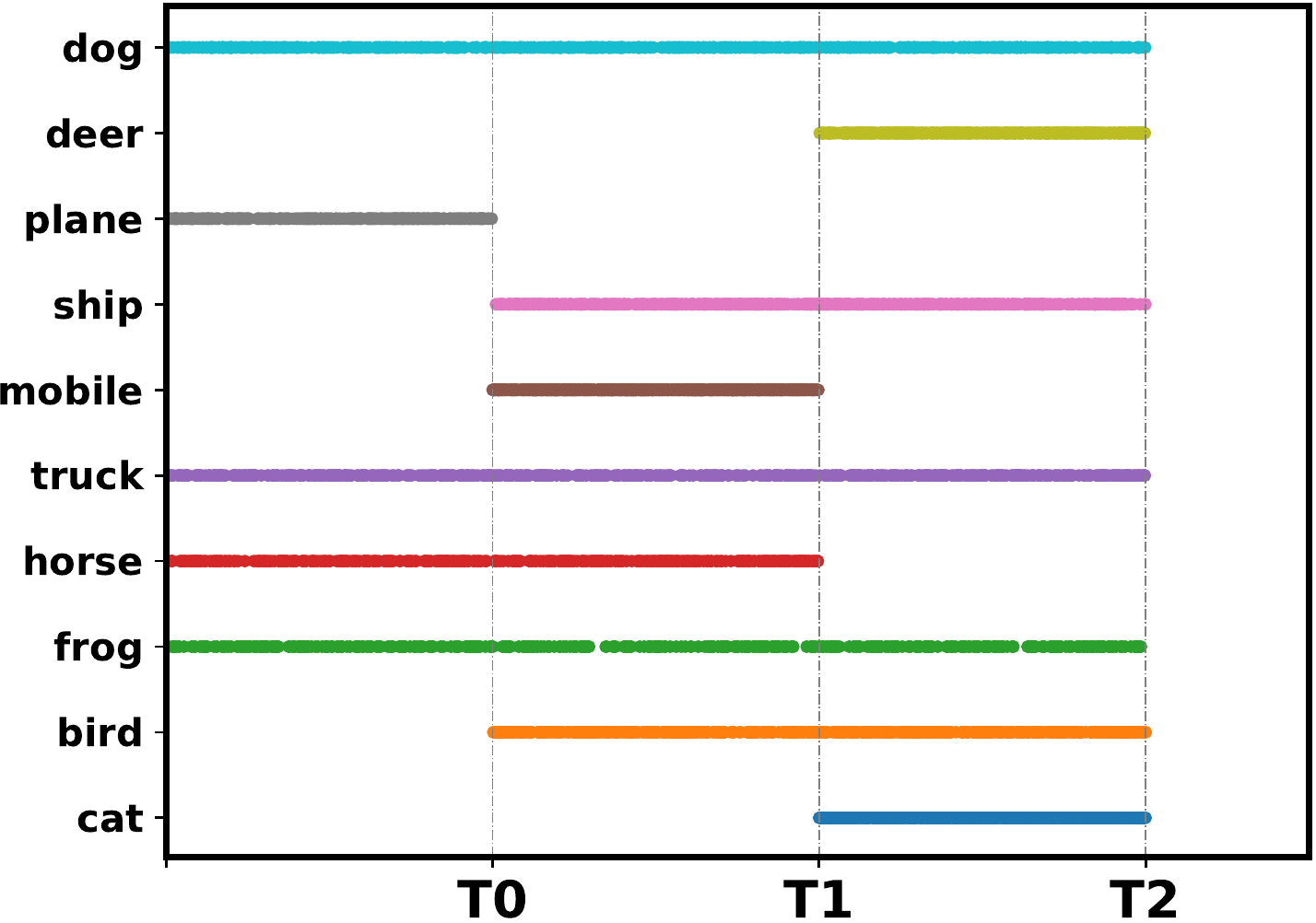}
			\mbox{  ({\it b}) {multiple unknown classes}}
		\end{minipage}
	\end{center}
	\caption{The class distribution of simulated streams on CIFAR-10 dataset. The X-axis denotes time scale and Y-axis is class information. (a) is streaming data with single unknown class, (b) denotes streaming data with multiple unknown classes. Note that some known classes may disappear in  time window $t$ in figure (b) in order to be more in line with real applications.}\label{fig:data}
\end{figure}

We consider single novel class case here following~\cite{DaYZ14,WangKCTK19,geng2020}. In detail, for each dataset, we randomly choose 50$\%$ from the total classes as known class set, the rests are regarded as unknown class set. The data of known class set can be divided into two parts: 1) 50$\%$ of data are regarded as initial training data; 2) the remaining data are used to constitute a streaming data. We simulate a streaming data as shown in Figure \ref{fig:data} (a). The data before time $t_0$ are training data. Then each class simulates an independent streaming data by shuffling instances randomly and arranging the data according to the index. A new class of streaming data appends every fixed time interval $\Delta t$. Thus, the instances occur in $\Delta t$ constitute a time window data mixed with known and unknown instances.
 
%Different from previous methods that only consider the classification at current time window, we consider a more comprehensive classification measure based on streaming data. In detail, we utilize 4 criteria, i.e., average Accuracy, average Recall, average Precision and average F1 to measure the performance. 

For calculating Forgetting criterion, let $acc_{k,j}$ be the accuracy evaluated on the known class set, i.e. the data of classes emerging on $j-$th time window ($j \leq k$), after training the network incrementally from stage $1$ to $k$, the average accuracy at time $k$ is defined as: $A_k = \frac{1}{k}\sum_{j=1}^k acc_{k,j}$ ~\cite{chaudhry2018riemannian}. Higher $A_k$ represents better classifier. Thus, to validate the catastrophic forgetting, we calculate the performance on forgetting profile as~\cite{chaudhry2018riemannian}, i.e., $Forgetting = \frac{A^{*} - mean(A)}{A^{*}}$, $A^{*}$ is the optimal accuracy with entire data, $A$ is the set of average accuracy.

\subsubsection{Incremental Multi-Class Case}\label{sec:s4}
In order to be more in line with the real scene, we consider additional case of multiple unknown classes here. The main difference from single unknown class is that more than one unknown class may appear in each time window. To solve this problem, considering that all datasets are with 10 classes, we firstly select 5 classes into known class set, the rest 5 classes are regarded as unknown class set. Then, we randomly select a number $l$ from 1-5 as the number of time windows, and randomly divide the unknown class set into $l$ parts. In result, we can obtain a incremental class order that appears in each time window. The generation of streaming data is the same as that of single unknown class setting. A case of generated streaming data of CIFAR-10 is shown in Figure \ref{fig:data} (b).

We only give the results of CIFAR-10 which is with complex content here. Table \ref{tab:tab4} compares the classification and forgetting performance of I-S2OSC with all baselines over streaming data under multiple novel classes case. Best results are in bold, N/A denotes no result. We can observe similar results as single novel class setting that I-S2OSC consistently outperforms all baselines on various criteria and has the least forgetting for update. We further compare the run time of I-S2OSC with other deep baselines. Considering the superior performance and model comparability, we only compare with CPE and DTC here. Specifically, the run time of I-S2OSC/CPE/DTC is 1.033/1.583/1.8 hours, we can observe that the run time of I-S2OSC is significantly less than other methods, because CPE and DTC employ losses based on embeddings, which convergences slowly.

\begin{table*}[htb]{
		\centering
		\caption{Comparison of incremental open set classification with multiple unknown classes.}
		\label{tab:tab4}
		\begin{tabular*}{1\textwidth}{@{\extracolsep{\fill}}@{}l|c|c|c|c|c|c|c|c|c}
			\toprule
			Methods & Iforest & One-SVM & LACU & SENC & ODIN & CFO & CPE  & DTC & I-S2OSC\\
			\midrule
			Accuracy  &.106 &.186 &.175  &.220 &.333 &.315 &.314  &.356 &\bf.444   \\
			Precision &.257 &.296 &.281  &.381 &.473 &.508 &.471  &.471 &\bf.606   \\
			Recall    &.106 &.186 &.175  &.220 &.333 &.315 &.356  &.356 &\bf.442   \\
			F1        &.132 &.205 &.194  &.134 &.271 &.259 &.399  &.399 &\bf.487   \\
			F$_{out}$ &.230 &.406 &.404  &.685 &.796 &.736 &.233  &.233 &\bf.802   \\
			Forgetting&N/A  &.625 &.633  &.678 &.606 &.390 &.374  &.374 &\bf.305  \\
			%			Time      &N/A  &. &.  &. &. &. &.  &. &\bf.  \\
			\bottomrule
	\end{tabular*}}
\end{table*}

\section{Broader Impact}
In this work, we study the problem of open set classification and incremental extension. We are aware that these technique may pose ethical issues, for example, people or institutions may use this technology to judge the efficiency of the system on detecting novel illegal pictures. Although there exist uncertainty within the technique, we consider that it is necessary to research and develop such techniques for understanding the dynamic environment. More reliable controls are expected from further discussion of these techniques. 

%In this work, we study the problem of automatic news comment generation, which is a novel explored task in the literature of natural language generation. We are aware that numerous uses of these techniques can pose ethical issues and that best practices will be necessary for guiding applications. In particular, we note that people expect comments on news to be made by people. Thus, there is a risk that people and organizations could use these techniques at scale to feign comments coming from people for purposes of political manipulation or persuasion. Also, we understand that the behaviors of deployed systems may need to be monitored and guided with methods, including post-processing techniques. While there are risks with this kind of AI research, we believe that developing and demonstrating such techniques is important for understanding valuable and potentially troubling applications of the technology. We hope to stimulate discussion about best practices and controls on these methods around responsible uses of the technology.

\bibliographystyle{nips}\small
\bibliography{supple}